\documentclass[10pt,journal,compsoc]{IEEEtran}
\ifCLASSOPTIONcompsoc
  \usepackage[nocompress]{cite}
\else
  \usepackage{cite}
\fi
\ifCLASSINFOpdf
  \usepackage[pdftex]{graphicx}
  \graphicspath{{../pdf/}{../jpeg/}}
\else
\fi
\usepackage{amsmath}
\usepackage{amssymb}
\usepackage{bbm}
\usepackage{bm}
\usepackage{soul}
\usepackage{xcolor}

\usepackage[utf8]{inputenc}
\usepackage[english, russian, french]{babel}

\usepackage{algorithm}
\usepackage{algpseudocode}
\usepackage{array}
\newcolumntype{M}[1]{>{\centering\arraybackslash}m{#1}}
\renewcommand{\arraystretch}{1.5}
\ifCLASSOPTIONcompsoc
  \usepackage[caption=false,font=footnotesize,labelfont=sf,textfont=sf]{subfig}
\else
  \usepackage[caption=false,font=footnotesize]{subfig}
\fi
\usepackage{float}

\usepackage{stfloats}
\usepackage{multirow}

\begin{document}
\selectlanguage{english}
	\title{Heterogeneous Domain Adaptation with Adversarial Neural Representation Learning: Experiments on E-Commerce and Cybersecurity}

\author{Mohammadreza~Ebrahimi,~\IEEEmembership{Member,~IEEE,}
        Yidong~Chai,
        Hao~Helen~Zhang,
        and~Hsinchun~Chen,~\IEEEmembership{Fellow,~IEEE}% <-this % stops a space

\IEEEcompsocitemizethanks
{\IEEEcompsocthanksitem M. Ebrahimi is with the School of Information Systems and Management, University of South Florida, 4202 E. Fowler Avenue, Tampa, FL 33620. E-mail: ebrahimim@usf.edu.\protect
\IEEEcompsocthanksitem  Y. Chai is with the School of Management, Hefei University of Technology. Corresspondence to: Y. Chai. Email: chaiyd@hfut.edu.cn.
\IEEEcompsocthanksitem Hao Helen Zhang is with the Department of Mathematics, University of Arizona. E-mail: hzhang@math.arizona.edu
\IEEEcompsocthanksitem Hsinchun Chen is with the Aritificial Intelligence Lab, University of Arizona. Email: hchen@eller.arizona.edu}% <-this % stops a space

}


\markboth{IEEE TRANSACTIONS ON PATTERN ANALYSIS AND MACHINE INTELLIGENCE, VOL. XX, NO. XX, XXXX 202X}%
{Shell \MakeLowercase{\textit{et al.}}: Bare Advanced Demo of IEEEtran.cls for IEEE Computer Society Journals}



\IEEEtitleabstractindextext{%
\selectlanguage{english}
\begin{abstract}
Learning predictive models in new domains with scarce training data is a growing challenge in modern supervised learning scenarios. This incentivizes developing domain adaptation methods that leverage the knowledge in known domains (source) and adapt to new domains (target) with a different probability distribution. This becomes more challenging when the source and target domains are in heterogeneous feature spaces, known as heterogeneous domain adaptation (HDA). While most HDA methods utilize mathematical optimization to map source and target data to a common space, they suffer from low transferability. Neural representations have proven to be more transferable; however, they are mainly designed for homogeneous environments. Drawing on the theory of domain adaptation, we propose a novel framework, Heterogeneous Adversarial Neural Domain Adaptation (HANDA), to effectively maximize the transferability in heterogeneous environments. HANDA conducts feature and distribution alignment in a unified neural network architecture and achieves domain invariance through adversarial kernel learning. Three experiments were conducted to evaluate the performance against the state-of-the-art HDA methods on major image and text e-commerce benchmarks. HANDA shows statistically significant improvement in predictive performance. The practical utility of HANDA was shown in real-world dark web online markets. HANDA is an important step towards successful domain adaptation in e-commerce applications.
\end{abstract}

\selectlanguage{english}
}

\maketitle

% is not selected <OR> if conference mode is selected - because compsoc
% conference papers position the abstract like regular (non-compsoc)
\IEEEdisplaynontitleabstractindextext

% \fi
\IEEEpeerreviewmaketitle

\selectlanguage{english}
\ifCLASSOPTIONcompsoc
\IEEEraisesectionheading{\section{Introduction}\label{sec:introduction}}
\else
\section{Introduction}
\label{sec:introduction}
\fi

\begin{figure}[!t]
\centering
\setlength\abovecaptionskip{-0.2\baselineskip}
\includegraphics[scale=0.5]{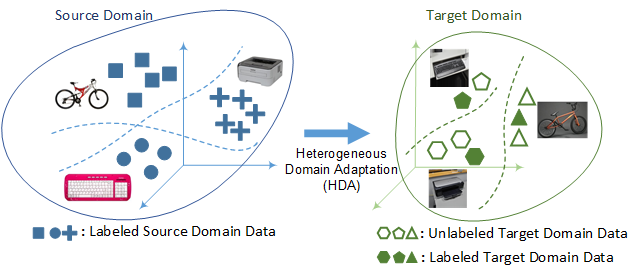}
%where an .eps filename suffix will be assumed under latex, 
%and a .pdf suffix will be assumed for pdflatex; or what has been declared
%via \DeclareGraphicsExtensions.
\caption{Illustrating HDA with Distribution Divergence and Feature Discrepancy. Three product categories are represented by different features and distributions in the source and target domains.}
\label{fig_intro}
\end{figure}

\IEEEPARstart{L}{earning} predictive models in new domains that lack enough training data has arisen as a challenge in supervised learning. This forms a strong motivation for transferring knowledge from common domains (source) to unknown domains (target), often known as Domain Adaptation (DA) \cite{cao_partial_2018,Ghifary2017,Zhong2020}. DA requires less supervision since it often operates with few labeled samples in the target domain. A practical example would be utilizing labeled user-generated content in legal e-commerce platforms (e.g., Amazon, eBay) to recognize unseen content in a new market and improve product search and indexing \cite{li_supervised_2018}. The same scenario applies to dark web e-commerce platforms (e.g., Dream Market and Russian Silk Road) in cybersecurity applications.\\
DA is a branch of transfer learning (TL) addressing two major issues: distribution divergence and feature discrepancy \cite{li_heterogeneous_2018}. The former arises because source domain samples admit a different distribution than that of the target domain. The latter occurs when source and target samples are expressed in different feature spaces. Most studies focus on the first issue, in which data distributions of source and target are different while samples are in the same feature space. Also known as homogeneous domain adaptation, this approach is not sufficient to address real-world scenarios.
Addressing both issues is significantly challenging and has emerged as a new field called Heterogeneous Domain Adaptation (HDA). In most HDA scenarios, labeled datasets are available from a known source, while the target domain suffers from a lack of labeled data (Figure \ref{fig_intro})
\cite{hoffman_cycada:_2018,li_deep_2018}.

%I wish you the best of success.

 

%\IEEEpubidadjcol

% An example of a floating figure using the graphicx package.
% Note that \label must occur AFTER (or within) \caption.
% For figures, \caption should occur after the \includegraphics.
% Note that IEEEtran v1.7 and later has special internal code that
% is designed to preserve the operation of \label within \caption
% even when the captionsoff option is in effect. However, because
% of issues like this, it may be the safest practice to put all your
% \label just after \caption rather than within \caption{}.
%
% Reminder: the "draftcls" or "draftclsnofoot", not "draft", class
% option should be used if it is desired that the figures are to be
% displayed while in draft mode.
%

Most HDA methods use mathematical optimization (e.g., convex optimization) or linear methods (e.g., linear discriminant analysis) to find common underlying representations from source and target \cite{fang_discriminative_2018,li_heterogeneous_2018,yao_low-resolution_2019}. These representations could be less transferable for DA \cite{ren_factorized_2018}. Neural representations have shown promise in DA due to their ability to automatically extract transferable features \cite{long_transferable_2018}. Transferable features are intermediate representations that are domain invariant \cite{Ganin2016,tzeng_adversarial_2017} and, thus, instrumental to successful DA \cite{li_deep_2018}. Adversarial Neural Domain Adaptation (ANDA) is a promising direction to obtain domain-invariant representations \cite{tzeng_adversarial_2017}, which leverages a game theory-based learning scheme \cite{goodfellow_generative_2014} to obtain high-quality representations.
As a subcategory of DA, existing ANDA methods often address homogeneous domain adaptation \cite{li_heterogeneous_2018}. However, heterogeneous sources of user-generated content are very common on e-commerce platforms (e.g., multilingual text or product images with different representations). We present a novel ANDA framework to extract transferable domain-invariant representations from heterogeneous domains. This framework extends ANDA to heterogeneous domains by employing an inverse problem-solving method known as dictionary learning to mitigate the feature discrepancy. It also alleviates the distribution divergence between source and target in an adversarial manner. Our proposed method significantly improves the performance of text classification and image recognition in e-commerce applications.
\selectlanguage{english}
\section{Related Work}
DA is a special case of TL. The goal of TL is to use the knowledge obtained from a resource-rich domain or task (i.e., source) to help a task or domain with insufficient training data (i.e., target) \cite{Weiss2016}. The domain consists of feature space and data probability distribution \cite{Pan2010}. The task is a function that can be learned on a domain to map the data to a corresponding label space. In many practical cases, the source and target tasks are often the same, while the domains vary. Hence, the source domain model often needs to be adapted to the target domain. For instance, in cybersecurity applications, a task could be determining if a dark net market product is cybersecurity-related when the domains are English (source) and Russian (target) product descriptions in the dark web. DA aims to transfer the knowledge to perform the same task in different source and target domains \cite{bengio_representation_2013}. This requires reducing the distribution divergence and feature discrepancy in the source and target domains \cite{bengio_representation_2013,li_heterogeneous_2018}. Most DA methods only focus on alleviating the distribution divergence (i.e., homogeneous DA) \cite{long_transferable_2018,long_transfer_2013,pan_domain_2011,wang_flexible_2014}. Homogenous DA does not apply to heterogeneous domains where the data dimensions and features are different. In contrast, Heterogeneous DA (HDA) is intended to reduce both feature discrepancy and distribution divergence.

\subsection{Heterogeneous Domain Adaptation (HDA)}\label{sec_hda}
HDA aims to perform the adaptation when the source and target domains are in different feature spaces. HDA methods often map features in the source and target to a common space using mathematical objective function optimization or a linear projection from source to target domain \cite{fang_discriminative_2018,hsieh_recognizing_2016,li_heterogeneous_2018,yao_low-resolution_2019,herath_learning_2017,tsai_learning_2016,xiao_feature_2015,zhou_heterogeneous_2014,li_learning_2014,hoffman_efficient_2013}. These features tend to be less transferable than representations obtained from neural networks for domain adaptation \cite{chen_transfer_2016,wang_heterogeneous_2018}. A taxonomy of selected major HDA methods is provided in Table \ref{table_hda}.

\begin{table*}[!t]
\setlength\abovecaptionskip{-0.2\baselineskip}
\setlength{\textfloatsep}{10pt plus 1.0pt minus 2.0pt}
\centering
\caption{A Taxonomy of Major HDA Methods}
\label{table_hda}
\begin{tabular}{|M{0.25\linewidth}|M{0.25\linewidth}|M{0.25\linewidth}|M{0.25\linewidth}|M{0.2\linewidth}|}
\hline
\textbf{Method} & \textbf{Method Category} & \textbf{Task} & \textbf{Testbed}\\
\hline
Discriminative Joint Distribution Adaptation (DJDA) \cite{yao_low-resolution_2019}
 & Objective function optimization & Low-resolution image recognition & Face images \\
\hline
HDA Network based on Autoencoder (HDANA) \cite{wang_heterogeneous_2018}  & Stacked autoencoder & Image recognition & Newswire articles, product images  \\
\hline
Progressive Alignment (PA) \cite{li_heterogeneous_2018}  & Objective function optimization & Multilingual text and image recognition & Newswire articles, product images  \\
\hline
Cross-Domain Mapping (CDM) \cite{fang_discriminative_2018} & Linear discriminant analysis & Multilingual text and image recognition & Newswire articles, product images\\
\hline
Invariant Latent Space (ILS) \cite{herath_learning_2017} & Kernel matching and optimization & Image and face recognition & Product and face images  \\
\hline
Cross-Domain Landmark Selection (CDLS) \cite{tsai_learning_2016}  & Objective function optimization & Image recognition & Product images  \\
\hline
Transfer Neural Trees (TNT) \cite{chen_transfer_2016} & Neural decision forest & Image recognition & Product images  \\
\hline
Generalized Joint Distribution Adaptation (G-JDA) \cite{hsieh_recognizing_2016}  & Objective function optimization & Image recognition and multilingual text classification & Newswire articles, product images  \\
\hline
Supervised Kernel Matching (SSKM) \cite{xiao_feature_2015}  & Kernel matching and optimization & Text and sentiment classification & Product reviews, newswire articles  \\
\hline
Sparse Heterogeneous Feature Representation (SHFR) \cite{zhou_heterogeneous_2014} & Objective function optimization & Text and sentiment classification & Newswire articles  \\
\hline
Semi-supervised Heterogeneous Feature Augmentation (SHFA) \cite{li_learning_2014} & Convex function optimization & Text and sentiment classification, Image recognition & Product reviews, newswire articles \\
\hline
Max-Margin Domain Transforms (MMDT) \cite{hoffman_efficient_2013} & Objective function optimization & Image recognition & Product images  \\
\hline
\end{tabular}
\end{table*}

As seen in the taxonomy, most HDA approaches obtain feature representations based on mathematical optimization and linear mapping, and few studies utilize neural network-based representations to accomplish HDA. Most extant methods cast the HDA objective into an optimization problem aimed at mapping source and target to a common feature space without promoting domain invariance between source and target, resulting in limited improvement in the transferability of the source model to the target domain. Next, we review the effective methods to attain domain invariance and position our study within the body of the related work.

\selectlanguage{english}
\subsection{Adversarial Neural Domain Adaptation (ANDA)}
One promising direction to achieve domain invariance in DA is to map source and target samples into a ‘common latent space,’ in which source and target representations of the same class are close to each other \cite{Chen2012,long_deep_2016}. Recently, adversarial learning has shown promise in providing domain-invariant representations without ‘pair alignment,’ eliminating the need for manually establishing sample correspondence in source and target \cite{hoffman_cycada:_2018,ren_factorized_2018}. Adversarial learning is a game-theoretic approach for simultaneously training two competing components, generator and discriminator \cite{Goodfellow2016}, which yields state-of-the-art performance in homogeneous DA \cite{long_transferable_2018}. When utilized in DA, the generator is trained to create representations that mimic both source and target distribution, and the discriminator is tasked to distinguish the source data from the target data created by the generator. ANDA generates adversarially learned domain-invariant representations from source and target that are suitable for domain adaptation \cite{tzeng_adversarial_2017}. These representations are useful in tasks such as text classification \cite{Ganin2016,liu_adversarial_2017,zhao_adversarial_2018}, character recognition \cite{hoffman_cycada:_2018,li_deep_2018,luo_label_2017,ren_factorized_2018,russo_source_2018}, or image classification \cite{liu_coupled_2016,long_transferable_2018,sankaranarayanan_generate_2018,zhang_domain-symmetric_2019,tzeng_adversarial_2017} on a variety of datasets such as English Amazon reviews, handwritten digits, and face images. However, existing ANDA methods mostly operate on sources and targets with homogeneous feature spaces and do not address heterogeneous domain adaptation. This restricts the applicability of DA in real-world applications where the source and target domains are both relevant to a task but do not share the same feature representation.

\selectlanguage{english}
\subsection{Research Positioning}
While ANDA can address the transferability issue, most ANDA frameworks consider homogeneous feature spaces. Given the heterogeneity of user-generated content (e.g., multilingual text, diverse image representations), these issues undermine the applicability of DA to emerging applications such as e-commerece or cybersecurity. To highlight the position of our proposed method, we categorize major DA work in two dimensions (Table {\ref{table_positioning}}). The horizontal dimension differentiates DA methods supporting heterogeneous features. The vertical dimension distinguishes between DA methods that generate neural representations and the others. Among homogeneous studies, methods relying on mathematical optimization include Support and Model Shift (SMS) \cite{wang_flexible_2014}, Joint Distribution Adaptation (JDA) \cite{long_transfer_2013}, and Transfer Component Analysis (TCA) \cite{pan_domain_2011}. Recent homogeneous methods such as Symmetric Networks (SymNets) \cite{zhang_domain-symmetric_2019}, Domain Adaptation Network (DAN) \cite{long_transferable_2018}, Symmetric Bi-directional ADAptive GAN (SBADA-GAN) \cite{russo_source_2018}, Auxiliary Classifier GAN (AC-GAN) \cite{sankaranarayanan_generate_2018}, Adversarial Discriminative Domain Adaptation (ADDA), and Cycle Consistent Adversarial Domain Adaptation (CYCADA) \cite{hoffman_cycada:_2018} support neural representations (the bottom left quadrant). Heterogeneous methods (the top right and bottom right quadrants) were recognized in Section {\ref{sec_hda}}. Little has been done to provide a solution that yields neural representations from heterogeneous domains (the bottom right quadrant).

\begin{table}[!b]
\setlength\abovecaptionskip{-0.2\baselineskip}
%% increase table row spacing, adjust to taste
\renewcommand{\arraystretch}{1.3}
% if using array.sty, it might be a good idea to tweak the value of
%\extrarowheight as needed to properly center the text within the cells
\caption{Positioning Our Proposed Method in DA Research}
\label{table_positioning}
\centering
%% Some packages, such as MDW tools, offer better commands for making tables
%% than the plain LaTeX2e tabular which is used here.
\begin{tabular}{|M{0.2\linewidth}|M{0.35\linewidth}|M{0.28\linewidth}|}
\hline
& \textbf{Homogeneous} & \textbf{Heterogeneous}\\
\hline
\textbf{Mathematical Optimization} & SMS \cite{wang_flexible_2014}, JDA \cite{long_transfer_2013}, TCA \cite{pan_domain_2011} & PA \cite{li_heterogeneous_2018}, CDM \cite{fang_discriminative_2018}, G-JDA \cite{hsieh_recognizing_2016}, CDLS \cite{tsai_learning_2016}, SHFA \cite{li_learning_2014}, MMDT \cite{hoffman_efficient_2013}\\
\hline
\textbf{Neural Representation} & SymNets \cite{zhang_domain-symmetric_2019}, DAN \cite{long_transferable_2018}, CYCADA \cite{hoffman_cycada:_2018}, AC-GAN \cite{sankaranarayanan_generate_2018}, SBADA-GAN \cite{russo_source_2018}, ADDA \cite{tzeng_adversarial_2017}& HDANA \cite{wang_heterogeneous_2018}, TNT \cite{chen_transfer_2016}, Our proposed model \\
\hline
\end{tabular}
\end{table}
Two extant neural representation-based HDA methods are recognized as alternatives for our proposed model: Transfer Neural Trees (TNT) \cite{chen_transfer_2016} and Heterogeneous Domain Adaptation Network based on Autoencoder (HDANA) \cite{wang_heterogeneous_2018}. TNT is a tree-based neural network architecture focused on conducting HDA via obtaining corresponding pairs in source and target domains and does not offer distribution alignment. HDANA is a deep autoencoder architecture addressing the distribution alignment with a fixed distribution divergence kernel, which may result in lack of domain invariance. Both these issues can lead to performance loss. To remedy these issues, our model introduces a novel HDA framework that enables both feature and distribution alignment in a unified neural network architecture with enhanced domain invariance. It utilizes dictionary learning and nonparametric adversarial kernel matching to achieve these goals without relying on a fixed kernel for distribution matching. Our model contributes to obtaining high-quality representations from heterogeneous domains, which are useful in downstream tasks such as multilingual text classification and image recognition.

\selectlanguage{english}
\section{Background for Proposed Model}
\selectlanguage{english}
\subsection{Model Preliminaries}
Let $\mathcal{D}^s=\{ \mathcal{X}^s,P(X^s,y^s)\}$  and $\mathcal{D}^t =\{ \mathcal{X}^t,Q(X^t,y^t)\}$ denote the source and target domains, each comprises feature spaces $\mathcal{X}^s$ and $\mathcal{X}^t$  and distributions $P$, $Q$ over the source and target samples, respectively. The source data $X^s=\{x_i^s,y_i^s\}_{i=1}^{n_s}$ includes labeled data sampled from distribution $P$. The target data $X^t=\{x_i^t\}_{i=1}^{n_t}$ is sampled from distribution $Q$ and contains both labeled data $\{(x_i^L,y_i^L)\}^{n_L}_{i=1} \sim {Q}$ and unlabeled data $\{(x_i^U)\}^{n_U}_{i=1} \sim {Q}$. HDA aims to help a classification task in the target domain by using the source domain data when not only is $P\ne{Q}$, but $\mathcal{X}^s\ne{\mathcal{X}^t}$. To achieve this, we present a neural network architecture to approximate a hypothesis $h(x)$ that not only minimizes the discrepancy between \(X^s\) and \(X^t\) as well as the divergence between distributions $P$ and $Q$, but also minimizes the error of assigning target labels $\epsilon_t(h(x))$, using source domain data and only a small number of labeled target data.
DA models need to be generalizable to unseen domains in order to reduce the target generalization error $\epsilon_t(h)$ \cite{long_transferable_2018}. Accordingly, we draw on the domain adaptation theory to identify the generalization error bound of the target \cite{ben-david_theory_2010,ben-david_analysis_2007} and inform the components of our proposed model. For any hypothesis space $\mathcal{H}$ and a fixed representation function the expected target error is bounded by Theorem 1.\\

\textbf{Theorem 1 \cite{ben-david_analysis_2007}:} Let ${X}^s$ and ${X}^t$ represent samples drawn from $P$ and $Q$, the source and target domain distributions, respectively. For any $\delta\in(0, 1)$, with probability at least $1-\delta$ (over the choice of the samples), for every hypothesis $h\in\mathcal{H}$:

\begin{equation}
\epsilon_t (h) \leq {\hat{\epsilon}}_s(h) + \hat{d}_\mathcal{H} ({X}^s,{X}^t )+\epsilon_s (h^*)+\epsilon_t (h^*)+C     
\label{eq_1}
\end{equation}	

where ${\hat{\epsilon}}_s (h)$ is the empirical source training error, $\epsilon_s (h^* )+\epsilon_t (h^* )$ is the combined error of the ideal hypothesis in both domains, and $C$ is a constant given in (\ref{eq_c}):

\begin{equation}
\begin{aligned}
C = \frac{4}{m}\sqrt{d\log\left(\frac{2em}{d}\right)+\log\left(\frac{4}{\delta}\right)} + 4\sqrt{\frac{d\log(2m')+log(\frac{4}{\delta})}{m'}}
\label{eq_c}
\end{aligned}
\end{equation}

where $e$ is the base for the natural logarithm, $d$ is the VC-dimension of the hypothesis space, $m$ denotes labeled sample size in source, and $m'$ is the unlabeled sample size. As seen in (\ref{eq_c}), $C$ only depends on the properties of the hypothesis space (i.e., VC-dimension and sample sizes), rather than the choice of samples. Assuming that the combined error of the ideal hypothesis $\epsilon_s (h^* )+\epsilon_t (h^* )$ is small for a reasonable representation, the bound in Theorem 1 depends on the first and second terms in the RHS of (\ref{eq_1}). It is expected that an effective DA model reduces these two terms and, thus, the target generalization error $\epsilon_t (h)$. The first error term ${\hat{\epsilon}}_s (h)$ can be reduced by learning shared representations from source and target \cite{ben-david_analysis_2007}, which requires aligning feature representations when the domains are heterogeneous. The second term denotes the empirical $\mathcal{H}$-divergence between the source and target distributions \cite{ben-david_theory_2010}:

\begin{equation}
\begin{aligned}
\hat{d}_\mathcal{H}({X}^s,{X}^t )\triangleq 2(1-\underset{ {h\in{\mathcal{H}}} }{\text{min}} \{\frac{1}{m'}\sum_{{x}:h({x})=0}\mathbbm{1}[{x}\in{{X}^s} ]\\ +\frac{1}{m'}\sum_{{x}:h({x})=1}\mathbbm{1}[{x}\in{{X}^t} ] \})
\end{aligned}
\label{eq_2}
\end{equation}

where $\mathbbm{1}$ is an indicator function that returns 1 when its argument is true and 0 otherwise. Long et al. \cite{long_transferable_2018} show that the minimization term in (\ref{eq_2}) is bounded by the Maximum Mean Discrepancy (MMD) distance under kernel $k$ \cite{smola_hilbert_2007}. Thus, (\ref{eq_1}) can be re-written as:

\begin{equation}
\begin{aligned}
\epsilon_t (h)\leq{\hat{\epsilon}}_s (h)+2(1+D_k (X^s,X^t ))+\lambda+C     
\end{aligned}
\label{eq_3}
\end{equation}

where $D_k (.)$ denotes the MMD distance in a Reproducing Kernel Hilbert Space (RKHS) with kernel $k$ \cite{smola_hilbert_2007}. The second term in (\ref{eq_3}) can be reduced by distribution kernel matching, aiming to minimize the distance of the source and target distribution.

To minimize these two terms, next we discuss dictionary learning as a promising approach to reduce ${\hat{\epsilon}}_s(h)$, which addresses feature discrepancy by translating source and target features to a common space. Subsequently, to reduce the $\mathcal{H}$-divergence between the source and target, we discuss distribution kernel matching and adversarial kernel learning.

\selectlanguage{english}
\subsection{Dictionary Learning: Reducing $\bm{{\hat{\epsilon}}_s(h)}$ }
A viable approach to align representations can be devised based on dictionary learning \cite{li_heterogeneous_2018}. Dictionary learning is an inverse problem that aims to decompose a given data matrix $X$ into a dictionary space $D$ and a coefficient matrix $R$, which denotes the representation of $X$ under $D$,  by optimizing (\ref{eq_4}):

\begin{equation}
\begin{aligned}
\underset{D,R}{\text{min}}\left\| X-DR\right\|_F^2 + \lambda\left\|R\right\|_1;\\
\text{s.t.} \left\|d_i\right\|_2\leq 1; \forall i \in 1,\ldots,k
\end{aligned}
\label{eq_4}
\end{equation}

where $F$ denotes the Frobenius norm measuring reconstruction error and $d_i$ is the $i^th$ column in $D$. Unlike other matrix factorization methods dictionary learning can promote sparsity in the coefficient matrix, resulting in extracting only salient features from the input. Dictionary learning has been successfully applied to signal \cite{garcia-cardona_convolutional_2018} and image \cite{li_heterogeneous_2018} processing, contributing to reducing the dimensionality of the input data while preserving the salient features. Sharing the dictionary $D$ between the source ($X^s$) and the target domain ($X^t$) forces the new representations to be projected to 
] space \cite{li_heterogeneous_2018}.

\begin{equation}
\begin{aligned}
&\underset{D,R^s,R^t}{\text{min}}\left\| X^s-DR^s\right\|_F^2 + \left\| X^t-DR^t\right\|_F^2 + \lambda\left\|R\right\|_1\\
&\text{s.t.} \left\|d_i\right\|_2\leq 1; \quad R=[R^s,R^t]
\end{aligned}
\label{eq_5}
\end{equation}

While this is useful in DA for aligning features, traditional dictionary learning ignores label information since it is not naturally designed for classification tasks often encountered in e-commerce applications. We are motivated to incorporate the dictionary learning objective function in a neural network architecture that can jointly learn a classification task such that it is possible to take advantage of labeled data in both the source and target domains.

\selectlanguage{english}
\subsection{Distribution Kernel Matching and Adversarial Kernel Learning: Reducing $\bm{\hat{d}_\mathcal{H}({X}^s,{X}^t )$}}

DA requires measuring the distribution divergence between source and target. This could be successfully achieved by defining a two-sample test statistic on source and target \cite{li_mmd_2017}. Two types of test statistics for DA are parametric and nonparametric. Parametric statistics measure the similarity of the density functions in source and target (e.g., Kullback-Leibler divergence) \cite{glorot_domain_2011,titov_domain_2011}. Nonparametric statistics measure the distance of kernels rather than the actual densities. This requires the distributions to be treated as functions in RKHS \cite{smola_hilbert_2007}. Nonparametric distribution kernel matching methods are often preferred for DA since they circumvent intractable density estimation \cite{long_transferable_2018}. Among nonparametric kernel matching methods, Maximum Mean Discrepancy (MMD) is the most widely used for DA \cite{chen_domain_2018,wang_heterogeneous_2018}. MMD measures the divergence based on the distance ${\rm D}$ between source and target domain data $X^s\sim{P}$ and $X^t\sim{Q}$.

\begin{equation}
\begin{aligned}
{\rm D}_k(X^s,X^t) \triangleq \left\| \mathbb{E}_{x^s}[\phi_{k}(x^s)] -  \mathbb{E}_{x^t}[\phi_{k}(x^t)] \right\|_{\mathcal{H}_k}^2 
	\end{aligned}
	\label{eq_6}
	\end{equation}

where $\phi_k (.)$ is a feature mapping from inputs to an embedding in the Hilbert space $\mathcal{H}_k$  characterized by kernel $k$. MMD’s test power is extremely sensitive to the choice of mapping $\phi_k$ \cite{gretton_kernel_2012,sriperumbudur_kernel_2009}, which could negatively affect the distribution alignment performance. Adversarial kernel learning can address this issue by facilitating finding the proper kernel that maximizes the test power of MMD \cite{li_mmd_2017}. Such a kernel helps align the distributions in source and target without establishing sample correspondence. To this end, the mapping $\phi$ in MMD can be optimized via adversarial kernel learning \cite{li_mmd_2017}. However, extant adversarial kernel learning methods are not designed for heterogeneous domains. Inspired by the recent advances, we employ adversarial kernel learning for HDA. This enables benefiting from the nonparametric property of MMD while improving the test power via adversarial learning.

\selectlanguage{english}
\section{Proposed Model: Heterogenous Adversarial Neural Domain Adaptation (HANDA)}
\begin{figure*}[!ht]
	\centering
	\setlength\abovecaptionskip{-0.2\baselineskip}
	\includegraphics[scale=0.6]{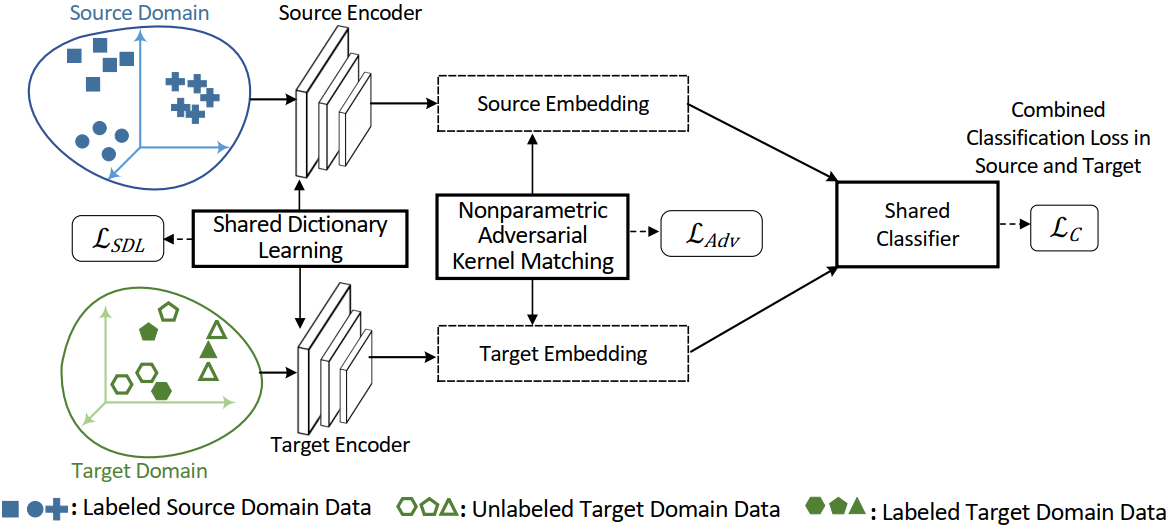}
	\caption{HANDA Architecture. The Shared Dictionary Learning (left) and Nonparametric Adversarial Kernel Matching (middle) components operate on the heterogeneous representations of source and target domain to reduce feature discrepancy via minimizing $\mathcal{L}_{SDL}$ and to reduce distribution divergence via minimizing $\mathcal{L}_{Adv}$, simultaneously. The Shared Classifier (right) component operates on the resultant representations via minimizing the shared classification loss $\mathcal{L}_{C}$.}
	\label{fig_model}
\end{figure*}

Informed by the effect of ${\hat{\epsilon}}_s(h)$ and $\hat{d}_\mathcal{H}(\tilde{X}^s,\tilde{X}^t )$ on reducing $\epsilon_t (h)$, we incorporate dictionary learning and adversarial kernel matching in a unified architecture that reduces the generalization error bound on any given domain adaptation problem. Accordingly, our proposed model integrates three components, each targeting an adaptation challenge in HDA. Figure \ref{fig_model} shows the integration of these three components in our model, Heterogeneous Adversarial Neural Domain Adaptation (HANDA). The first component is a shared dictionary learning (SDL) approach that targets alleviating feature discrepancy via projecting heterogeneous features into a common latent space and is associated with minimizing a reconstruction loss $\mathcal{L}_{SDL}$. The second component is a nonparametric adversarial kernel matching method, which aims to reduce the distribution divergence by combining the nonparametric benefits of MMD and flexibility of kernel choices in adversarial learning. This component is associated with minimizing an adversarial loss denoted by $\mathcal{L}_{Adv}$. Finally, the third HANDA’s component is a shared classifier in source and target that targets limited labeled data in the target domain by exploiting labeled samples in both the source and target domain. This component yields the classifier $h(x)$ as the ultimate goal of HDA and is associated with minimizing the classification loss $\mathcal{L}_C$.

While shared dictionary learning aligns heterogeneous features in a common latent space and yields a new embedding, nonparametric adversarial kernel matching reduces the distribution divergence in the obtained embedding. After feature and distribution alignment, a shared classifier conducts the downstream tasks (e.g., text or image classification). HANDA’s novelty is threefold. First, it extends ANDA to heterogeneous domains. Second, it enhances dictionary learning to take advantage of labeled data in the source and target, and third, it enables joint feature and distribution alignment in a unified neural network architecture. We describe each HANDA component next.

\subsection{Component \#1: Shared Dictionary Learning (SDL)}
To minimize feature discrepancy, the source and target need to be mapped into the same subspace by domain-specific projections $P^s$ and $P^t$. This can be achieved by modifying (\ref{eq_5}) to get:

\begin{equation}
\begin{aligned}
\underset{P^s,P^t,D,R^s,R^t}{\text{min}} \left\| P^sX^s-DR^s\right\|_F^2
+ &\left\| P^tX^t-DR^t\right\|_F^2 \\
+ \lambda\left\|R\right\|_1;\\
\text{s.t.} \left\|d_i\right\|_2\leq 1; \quad R=[R^s,R^t]
\end{aligned}
\label{eq_7}
\end{equation}

where $P_{k \times m_s}^s$ and $P_{k \times m_t}^t$ are projections. $X_{m_s \times n_s}^s$ and $X_{m_t \times n_t}^t$ are the source and target data. $D_{k \times k}$ is the shared dictionary, and $R_{k \times n_s}^s$ and $R_{k \times n_t}^t$ denote the aligned representations. To avoid degenerate solutions, $P^s$ and $P^t$ are often constrained to be orthogonal \cite{shekhar_generalized_2013}. The minimization in (7) yields data representations $R^s$ and $R^t$ that are in the same feature space.

%\begin{figure}[H]
%	\centering
%	\includegraphics[width=2.3in]{fig_mapping}
%	\caption{Mapping Source and Target to a Common Feature Space.}
%	\label{fig_mapping}
%\end{figure}

\subsection{Component \#2: Nonparametric Adversarial Kernel Matching}
Even though the obtained representations $R^s$ and $R^t$ are in the same space, their distributions are different. Thus, minimizing distribution divergence is crucial for successful DA. MMD measures the distribution divergence of source and target via a ‘two-sample test,’ in which the source and target distributions are the same under the null hypothesis. The distribution divergence can be minimized by learning a model $\phi_\theta (.)$ parameterized by $\theta$ that minimizes the MMD distance:

\begin{equation}
\begin{aligned}
\underset{\theta}{\text{min}} \quad {\rm D}_k(\phi_k(\phi_\theta(R^s)), \phi_k(\phi_\theta(R^t)))\\
\end{aligned}
\label{eq_8}
\end{equation}

where ${\rm D}_k (.)$ is the MMD distance with a mapping function of some kernel $k$. The explicit computation can be avoided by applying the kernel trick $k(x,x' )= \langle\phi(x),\phi(x')\rangle_\mathcal{H}$ \cite{gretton_kernel_2012}, which gives:

\begin{equation}
\begin{aligned}
{\rm D}_k(X^s,X^t) \triangleq \mathbb{E}_{x^s,x'^s}[k(x^s,x'^s)]&-2\mathbb{E}_{x^s,x^t}[k(x^s,x^t)]\\
& + \mathbb{E}_{x^t,x'^t}[k(x^t,x'^t)]
\end{aligned}
\label{eq_9}
\end{equation}

The test power of $D_k (.)$ in (9) heavily relies on the choice of the kernel $k$ \cite{li_mmd_2017}. Low test power can preclude correctly distinguishing unlike source and target domains. To increase the test power in DA, it is natural to search for the kernel that maximizes the distance between source and target. Thus, (\ref{eq_8}) can be re-written as:

\begin{equation}
\begin{aligned}
\underset{\theta}{\text{min}} \quad \underset{k}{\text{max}} \quad {\rm D}_k(\phi_k(\phi_\theta(R^s)), \phi_k(\phi_\theta(R^t)))\\
\end{aligned}
\label{eq_10}
\end{equation}

To cover a wide range of kernels (as opposed to a fixed one), we parameterize the kernel using neural network $M$ and search for the optimal kernel by learning the network parameters:

\begin{equation}
\begin{aligned}
\underset{\theta}{\text{min}} \quad \underset{\theta_M}{\text{max}} \quad 
{\rm D}_{\theta_M}(\phi_{\theta_M}(\phi_\theta(R^s)), \phi_{\theta_M}(\phi_\theta(R^t)))\\
\end{aligned}
\label{eq_11}
\end{equation}

Inspired by the idea of adversarial kernel learning \cite{li_mmd_2017}, we suggest minimizing (\ref{eq_11}) via another neural network $N$ in a GAN-like fashion \cite{goodfellow_generative_2014}. These two neural networks can be trained in an adversarial manner to solve the minimax problem in (\ref{eq_12}).

\begin{equation}
\begin{aligned}
\underset{\theta_N}{\text{min}} \quad \underset{\theta_M}{\text{max}} \quad {\rm D}_{\theta_M}(\phi_{\theta_M}(\phi_{\theta_N}(R^s)), \phi_{\theta_M}(\phi_{\theta_N}(R^t)))\\
\end{aligned}
\label{eq_12}
\end{equation}

\begin{figure*}[!t]
	\centering
	\setlength\abovecaptionskip{-0.2\baselineskip}
	\includegraphics[scale=0.6]{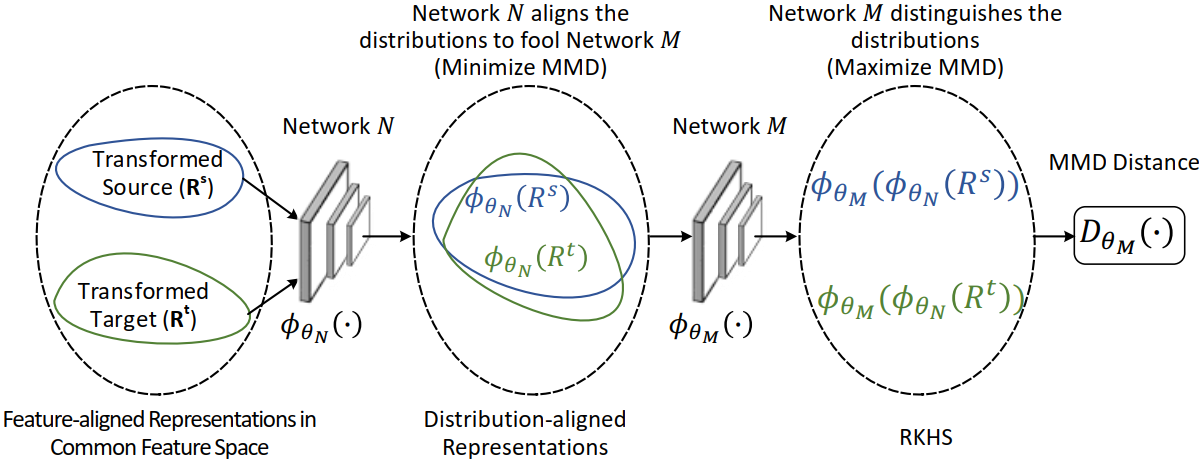}
	%where an .eps filename suffix will be assumed under latex, 
	%and a .pdf suffix will be assumed for pdflatex; or what has been declared
	%via \DeclareGraphicsExtensions.
	\caption{Abstract View of Adversarial Kernel Matching. Network $N$ aligns the distributions to fool Network $M$ via minimizing MMD between the source and target representations while network $M$ aims to yield the kernel with maximal test power between two distributions via maximizing MMD.}
	\label{fig_minimax}
\end{figure*}

Figure \ref{fig_minimax} illustrates the abstract view of the adversarial game defined in (\ref{eq_12}).

Transformed features obtained from dictionary learning in the source and target domains participate in an adversarial setup in which they are aligned via network $N$ to minimize the MMD distance between the domains, while network $M$ aims to maximize MMD, searching for the kernel with maximal test power.

\subsection{Component \#3: Shared Classifier for Source and Target}
With the aligned features and distributions, a shared neural network $h_C$ can be applied to output the class probability of each sample in the source and target, respectively. Given the suitability of hinge loss \cite{cai_multi-class_2011} in multi-class problems, the source and target classification losses are defined as $\mathcal{L}_s=L(h_C (\phi_{\theta_N} (R^s )),Y^s)$ and  $\mathcal{L}_t=L(h_C (\phi_{\theta_N} (R^t )),Y^t)$, where $L(.)$ denotes the multi-class hinge loss.

We obtain the objective function for the shared classification hypothesis $h_C$ by minimizing a convex combination of $L_s$ and $L_t$ \cite{ben-david_theory_2010,long_transferable_2018}, as given in (\ref{eq_13}):

\begin{equation}
\begin{aligned}
\underset{\theta_C}{\text{min}} \quad \frac{1}{2}(L(h_C (\phi_{\theta_N} (R^s )),Y^s)+L(h_C (\phi_{\theta_N} (R^t )),Y^t))\\
\end{aligned}
\label{eq_13}
\end{equation}

Finally, the total objective function of HANDA is obtained by aggregating the three losses in an adversarial manner: shared dictionary loss $(\mathcal{L}_{SDL})$ obtained from (\ref{eq_7}), MMD distance loss calculated via adversarial kernel matching $(\mathcal{L}_{Adv})$ from (\ref{eq_12}), and combined classification losses in source and target $(\mathcal{L}_C)$ from (\ref{eq_13}). Thus, the total objective function for our proposed method is given by (\ref{eq_14}):

\begin{equation}
\begin{aligned}
\underset{\theta_N,\theta_C,\Theta_{SDL}}{\text{min}} \quad \underset{\theta_M}{\text{max}} \quad \beta\mathcal{L}_{SDL}+ \gamma\mathcal{L}_{Adv} + \mathcal{L}_C\\
\end{aligned}
\label{eq_14}
\end{equation}

where $\Theta_{SDL} =\{P^s,P^t,D, R^s,R^t\}$ is the set of parameters for dictionary learning. $\theta_N$ and $\theta_M$ denote the parameters of adversarial networks for kernel matching, and $\theta_C$ is the classifier network’s parameters. $\beta$ and $\gamma$ are the trade-off parameters for the penalty incurred by corresponding losses. In Section 5, we demonstrate that these trade-off parameters can be empirically tuned via grid search.

\subsection {Learning Algorithm}
Past research has attempted to design DA algorithms that reduce target generalization error. Li et al. \cite{li_heterogeneous_2018} offer a representation learning algorithm based on objective function optimization with a ‘fixed’ MMD kernel without kernel learning. Long et al. \cite{long_transferable_2018} propose a deep representation learning algorithm with adversarial kernel learning. However, their algorithm is designed for homogeneous DA. As shown in model preliminaries, SDL and adversarial kernel learning can reduce HDA target generalization errors. Hence, we are motivated to incorporate SDL and nonparametric adversarial kernel matching in our algorithm to reduce the target error more effectively. We develop an algorithm to solve the optimization problems for SDL in (\ref{eq_7}), nonparametric adversarial kernel matching in (\ref{eq_12}), and classification in (\ref{eq_13}) via a unified architecture.
To achieve this, we incorporate SDL into our model. In the SDL formulation in (\ref{eq_7}), if $D$ is a low-rank matrix, the $\ell_1$ regularization term can be removed without lack of generalization:

\begin{equation}
\begin{aligned}
&\underset{P^s,P^t,D,R^s,R^t}{\text{min}} \left\| P^sX^s-DR^s\right\|_F^2
+ \left\| P^tX^t-DR^t\right\|_F^2 \\
&\text{s.t.} \left\|d_i\right\|_2\leq 1; \quad P^s{P^s}^T = I; P^t{P^t}^T=I
\end{aligned}
\label{eq_15}
\end{equation}

where $I$ denotes the identity matrix and the last two constraints ensure orthogonality. Ideally, the reconstruction errors are reasonably close to 0, and the source and target representations are expressed as ${\hat{R}}^s =D^{-1}P^s X^s$ and ${\hat{R}}^t=D^{-1}P^t X^t$. Accordingly, we express $R^s$ and $R^t$ in the form of products of three matrices where the first matrix is shared. Specifically, we denote $R^s\approxeq{AB^s X^s}$ and $R^t\approxeq{AB^t X^t}$, where $B^s$ and $B^t$ approximate the domain-specific projections and $A$ approximates the shared projection. To approximate $R^s$ and $R^t$, we parameterize $D^{-1}$, $P^s$, and $P^t$ by fully connected layers $A$, $B^s$, and $B^t$, respectively.

%\begin{figure}[H]
%	\centering
%	\includegraphics[scale=0.65]{SDL}
%	\caption{Abstract View of Shared Dictionary Learning.}
%	\label{fig_SDL}
%\end{figure}

Such approximation enables us to integrate shared dictionary learning into our neural network architecture. As a result, (\ref{eq_15}) can be written as:

\begin{equation}
\begin{aligned}
\underset{P^s,P^t,D,A,B^s,B^t}{\text{min}} \left\| P^sX^s-DAB^sX^s\right\|_F^2
+ &\left\| P^tX^t-DAB^tX^t\right\|_F^2 \\
\text{s.t.} \left\|d_i\right\|_2\leq 1; \quad P^s{P^s}^T = I; P^t{P^t}^T = I
\end{aligned}
\label{eq_16}
\end{equation}

To satisfy the orthogonality constraints on $P^s$ and $P^t$ in (\ref{eq_16}), we enforce the weights of $B^s$ and $B^t$ to be orthogonal after each gradient update. We also normalize the weights of $D$ and $A$ to the unit vector. Since $R^s$ and $R^t$ are restricted to be a function of $X^s$ and $X^t$, the result of the optimization in (\ref{eq_16}) is an upper bound for shared dictionary learning loss:

\begin{equation}
\begin{aligned}
\mathcal{L}_{SDL^{up}} = \left\| P^sX^s-DAB^sX^s\right\|_F^2
+ \left\| P^tX^t-DAB^tX^t\right\|_F^2 \\
\end{aligned}
\label{eq_17}
\end{equation}

Given the approximated representations of $R^s$ and $R^t$, the other loss functions in objective (\ref{eq_14}) are re-written as in (\ref{eq_18}) and (\ref{eq_19}):
%\vspace{-20pt}
\begin{equation}
\begin{aligned}
\mathcal{L}_{Adv} = D_{\theta_M}(\phi_{M}(\phi_{N}(AB^sX^s)), \phi_{M}(\phi_{N}(AB^tX^t)))\\
\end{aligned}
\label{eq_18}
\end{equation}

\begin{equation}
\begin{aligned}
\mathcal{L}_C = \frac{1}{2}( L(h_C (\phi_{\theta_N} (AB^sX^s )),Y^s)+\\L(h_C (\phi_{\theta_N} (AB^tX^t)),Y^t))
\end{aligned}
\label{eq_19}
\end{equation}

\begin{algorithm}[!h]
	\caption{HANDA Learning Procedure}
	\label{Algo}
	\begin{algorithmic}[1]
		\State	\textbf{Input}: trade-off parameters $\gamma$ and $\beta$; batch size for source data, target labeled data, and target unlabeled data: $b_s$, $b_L$, and $b_U$; \# of updates of dictionary learning $n_d$; \# of updates of adversarial kernel matching $n_a$; \# of updates of classification hinge loss $n_c$.		
		\State \textbf{Output}: Parameters $A$ ,$B^s$, $B^t$, $D$, $P^s$, $P^t$, $\theta_N$, $\theta_M$,  $\theta_C$.
		\State Initialize parameters $A,B^s,B^t,D,P^s,P^t,\theta_N,\theta_M$, $\theta_C$.
		\While{$\mathcal{L}_{SDL^{up}}$, $\mathcal{L}_{Adv}$, and $\mathcal{L}_{C}$ have not converged}
		\State Randomly sample a minibatch of source data
		\par 
		$\{(x_i^s,y_i^s)\}^{b_s}_{i=1}$.
		\State Randomly sample a minibatch of target data \par $\{(x_i^L,y_i^L)\}^{b_L}_{i=1}$ and $\{(x_i^U)\}^{b_U}_{i=1}$.
		\State Compute $R^s=AB^sX^s$ and $R^t=AB^tX^t$.
		%\State \textbackslash\textbackslash Update $\Theta_{SLD^up}$ for dictionary learning
		\For {$t = 1,2,...,n_{d}$}
		\State Compute shared dictionary learning loss $\mathcal{L}_{SDL}$\par 
		\hskip\algorithmicindent using (\ref{eq_17}).
		\State Update $A, B^s, B^t, D, P^s, P^t$ using stochastic \par 
		\hskip\algorithmicindent gradient descent to minimize $\mathcal{L}_{SDL}$.
		\State {\textbackslash\textbackslash Normalize weights to unit norm}
		\State $D \leftarrow $clip$(D, 1)$, $A \leftarrow $clip$(A, 1)$
		\State {\textbackslash\textbackslash Enforce orthogonality constraint}
		\State $P^s, P^t, B^s, B^t  \leftarrow $Orth$(P^s,P^t, B^s, B^t)$
		\EndFor
		\For {$t = 1,2,...,n_{a}$}
		\State Compute adversarial loss for distribution \par 
		\hskip\algorithmicindent alignment $\mathcal{L}_{Adv}$ using (\ref{eq_18}).
		\State Update parameters $\theta_M$ using stochastic gradient \par 
		\hskip\algorithmicindent descent to maximize $\mathcal{L}_{Adv}$.
		\State	Update parameters $\theta_N, A, B^s, B^t$ via stochastic \par 
		\hskip\algorithmicindent gradient descent to minimize $\mathcal{L}_{Adv}$.
		\EndFor
		\For {$t = 1,2,...,n_{c}$}
		\State Compute classification loss $\mathcal{L}_{C}$ using (\ref{eq_19}).
		\State Update $\theta_C,\theta_N,A, B^s, B^t, D, P^s, P^t$ using \par 
		\hskip\algorithmicindent stochastic gradient descent to minimize $\mathcal{L}_{C}$.
		\EndFor		
		\EndWhile
		
		\State \Return $A$ ,$B^s$, $B^t$, $D$, $P^s$, $P^t$, $\theta_N$, $\theta_M$,  $\theta_C$.
	\end{algorithmic}
\end{algorithm}

All three losses can be minimized simultaneously via stochastic gradient descent \cite{li_mmd_2017} in a unified neural network architecture introduced by HANDA. Algorithm 1 summarizes HANDA’s learning procedure to minimize these losses simultaneously. Algorithm 1 alternates between minimizing the feature discrepancy and minimizing the distribution divergence until some convergence criterion is met or for a number of iterations. Code and datasets are available at https://github.com/mohammadrezaebrahimi/HANDA. Our algorithm differs from the heterogeneous method proposed in \cite{li_heterogeneous_2018} in two aspects. First, we incorporate SDL in a unified neural network architecture that allows utilizing the supervised loss to improve the quality of dictionary learning. The dictionary learning in \cite{li_heterogeneous_2018} is unsupervised and does not account for the class variable in feature alignment, which could result in lack of discriminative properties in the projected space. Second, our proposed neural network architecture offers an effective alternative for the mathematical optimization based on Lagrange multipliers in \cite{li_heterogeneous_2018} via a numerical optimization (stochastic gradient-based) method that reduces distribution and feature mismatch simultaneously. Later in the evaluation section we show the performance benefit gained from this property. Our method also differs from the homogeneous method offered in \cite{tzeng_adversarial_2017} in that HANDA does not use GAN. While GAN implements a minimax setting between a generator and discriminator, the minimax optimization in our approach is formulated for adversarial kernel matching between source and target distributions, as given in (19), to achieve domain invariance.

\selectlanguage{english}
\section{Evaluation}
For a comprehensive evaluation, we measure both HANDA’s performance and its practical utility. We construct a large testbed with three e-commerce datasets. To evaluate the performance, we use two widely used e-commerce benchmarks for image and multilingual text: Reuters Multilingual Collection and Office31 – Caltech256. To evaluate the practical utility, we use a real-world e-commerce dataset from dark web online markets with multilingual product descriptions. This practical evaluation encompasses a real-world cybersecurity case study, which is of value for cybersecurity applications.
Reuters Multilingual Collection contains $111,740$ online newswire articles with commercial themes in five languages: English, French, German, Italian, and Spanish, across six topics: Economics, Equity Markets, Finance, Industry, Social, and Performance \cite{amini_learning_2009}. Consistent with other studies \cite{fang_discriminative_2018,tsai_learning_2016}, we used Spanish as the target language and the other four languages as sources. This dataset is used for performing multi-class text classification task. Office31 – Caltech256 is an extension of the original dataset created by Gong et al. \cite{gong_connecting_2013}, and contains two popular heterogeneous image representations: Speeded Up Robust Features (SURF) with $800$ dimensions \cite{fang_discriminative_2018} and Deep Convolutional Activation Features (DeCAF) with $4,096$ dimensions \cite{donahue_decaf:_2014}. The dataset consists of $2,533$ online images in 10 categories in four domains: Amazon (A), downloaded from the Amazon website; Webcam (W), taken by a low-resolution camera; DSLR (D), taken by a high-resolution DSLR camera; and Caltech(C), selected from the Caltech-256 dataset. Following other studies \cite{fang_discriminative_2018,tsai_learning_2016}, three labeled samples from each category within the source domain were used in training. This dataset is associated with multi-class product image recognition task. In addition to widely accepted benchmark e-commerce datasets, it is useful to evaluate the practical utility of our model in a real-world setting. To this end, given the recent rise of illegal online markets on the dark web, we developed crawlers for obtaining data from nine illegal e-commerce platforms. Our new dark web online markets dataset is an extension of the dataset proposed in \cite{ebrahimi_detecting_2018}, which contains {$94,175$} product descriptions (from seven English markets), {$4,600$} Russian, and {$1,512$} French product descriptions (Table \ref{table_darkweb}). The dataset has been labeled by three native speakers and two cybersecurity experts and is publicly available from our GitHub repository. To construct the training set, we randomly sampled product descriptions in each language (2,584 products; 706 cyber threats, 1,878 benign products). Since more English labeled data is available, English markets are used as the source, and Russian and French markets are used as the targets. Each language was manually labeled by a cybersecurity expert and a native speaker. This dataset is used for cyber threat detection, which is viewed as a binary classification task. 
\begin{table}[t]
	%% increase table row spacing, adjust to taste
	\renewcommand{\arraystretch}{1.3}
	\setlength\abovecaptionskip{-0.3\baselineskip}
	% if using array.sty, it might be a good idea to tweak the value of
	%\extrarowheight as needed to properly center the text within the cells
	\caption{Testbed for Cybersecurity Case Study }
	\label{table_darkweb}
	\centering
	%% Some packages, such as MDW tools, offer better commands for making tables
	%% than the plain LaTeX2e tabular which is used here.
	\begin{tabular}{|p{0.25\linewidth}|p{0.15\linewidth}|p{0.2\linewidth}|p{0.15\linewidth}|}
		\hline
		\multicolumn{1}{|c|}{Online Market} & \multicolumn{1}{c|}{\# of listings} & \multicolumn{1}{c|}{Language} & \multicolumn{1}{c|}{\# of labeled products}\\
		\hline
		Dream Market & 39,473 & \multirow{7}{*}{English} & \multirow{7}{*}{1,821}\\\cline{1-2}
		AlphaBay & 25,118 &  & \\\cline{1-2}
		Hansa & 14,149 &  & \\\cline{1-2}
		Silk Road 3 & 1,683 &  & \\\cline{1-2}
		Minerva & 683 &  &\\\cline{1-2}
		Apple Market & 877 &  & \\\cline{1-2}
		Valhalla & 12,192 &  & \\
		\hline
		Russian Silk Road & 4,600 & Russian & 552 \\
		\hline
		French Deep Web & 1,512 & French & 211\\
		\hline
		\textbf{Total:} & \textbf{100,287} & - & \textbf{2,584}\\
		\hline
		
	\end{tabular}
	\footnotetext{SMS: Support and Model Shift, JDA: Joint Distribution Adaptation, TCA: Transfer Component Analysis; SymNets: Symmetric Networks; SBADA-GAN: Symmetric Bi-directional ADAptive GAN; AC-GAN: Auxiliary Classifier GAN; DAN: Domain Adaptation Networks; CyCADA: Cycle Consistent Adversarial Domain Adaptation.}
\end{table}
The current body of HDA research widely uses average accuracy and standard deviation for performance evaluation \cite{li_heterogeneous_2018,wang_heterogeneous_2018,yao_low-resolution_2019}. The average accuracy $Acc_{avg}$ is defined as follows:

\begin{equation}
\begin{aligned}
Acc_{avg} =   \frac{1}{n} \sum_{i=1}^n \frac{| x^t:x^t\in X^t, {\hat{y}}^t=y^t |} {| x^t:x^t \in X^t |}
\end{aligned}
\label{eq_20}
\end{equation}

where $x$ is a target instance, ${\hat{y}}^t$ is the predicted class, $y^t$ is the true class label, $|.|$ denotes the cardinality of the corresponding set, and $n$ is the number of runs. Standard deviation is also averaged over multiple runs of each model (often 10 times). Consistent with past HDA research, we use average accuracy and average standard deviation as evaluation metrics for comparing HANDA with the benchmark methods. Higher average accuracy and lower standard deviation suggest higher predictive performance. Note that since both benchmark datasets are not class imbalanced, accuracy is a sufficient measure for benchmark evaluations. However, for class imbalanced cybersecurity datasets, the Area Under receiver operating characteristic Curve (AUC) is the recommended performance measure since it emphasizes both threat and non-threat classes \cite{wheelus_tackling_2018} by establishing a trade-off between Type I and Type II errors \cite{Hastie2017}. Given the class imbalance in our case study, we use AUC as the performance metric. Higher AUC suggests higher predictive performance.\\
We evaluate HANDA through three sets of experiments and one case study. Experiment \#1 is aimed at evaluating HANDA against alternative state-of-the-art HDA methods (the top right and bottom right quadrants, shown in Table \ref{table_positioning}). This experiment compares HANDA’s performance to seven HDA methods, including HDA Network based on Autoencoder (HDANA) \cite{wang_heterogeneous_2018}, Cross-Domain Mapping (CDM) \cite{fang_discriminative_2018}, Generalized Joint Distribution Adaptation (G-JDA) \cite{hsieh_recognizing_2016}, Transfer Neural Trees (TNT) \cite{chen_transfer_2016}, Cross-Domain Landmark Selection (CDLS) \cite{tsai_learning_2016}, Semi-supervised Heterogeneous Feature Augmentation (SHFA) \cite{li_learning_2014}, and Max-Margin Domain Transforms (MMDT) \cite{hoffman_efficient_2013}. This experiment involves two reputable e-commerce datasets: Office31 – Caltech256 and Reuters Multilingual Collection. Experiment \#2 focuses on the qualitative evaluation of DA \cite{chen_transfer_2016} to verify its effectiveness by comparing the linear separability of target samples before and after DA through visualizing the intermediate representations on both Office31 – Caltech256 and Reuters Multilingual Collection. Experiment \#3 targets convergence analysis \cite{arjovsky_wasserstein_2017} to investigate the convergence property of the objective functions corresponding to each HANDA component by assessing the stabilization of loss values during model training on the Office31 – Caltech256 benchmark dataset. Finally, a case study is conducted to verify practical utility on real-world products advertised in dark web online markets as an emerging cybersecurity application.

\begin{table*}[!t]
	\setlength\abovecaptionskip{-0.3\baselineskip}
	\centering
	\caption{Comparison of State-of-the-Art HDA Methods with HANDA on Multilingual Text Classification (Accuracy and Standard Deviation); P-values significant at 0.05:*}
	\label{table_result1}
	\begin{tabular}{|M{0.15\linewidth}|M{0.1\linewidth}|M{0.1\linewidth}|M{0.1\linewidth}|M{0.1\linewidth}|M{0.1\linewidth}|M{0.1\linewidth}|M{0.1\linewidth}|M{0.1\linewidth}|}
		\hline
		\multirow{2}{*}{HDA Method}& \multicolumn{4}{c|}{10 Labeled Samples per Target Class} & \multicolumn{4}{c|}{20 Labeled Samples per Target Class}\\\cline{2-9}
		& $EN\rightarrow SP$ & $FR\rightarrow SP$ & $GR\rightarrow SP$ & $IT\rightarrow SP$  & $EN\rightarrow SP$ & $FR\rightarrow SP$ & $GR\rightarrow SP$ & $IT\rightarrow SP$\\
		\hline
		MMDT & $68.9\pm0.6$ & $69.1\pm0.7$ & $68.3\pm0.6$ &	$67.5\pm0.5$ & $74.5\pm0.7$ &	$74.9\pm0.7$ &	$75.1\pm0.8$ &	$74.9\pm0.8$  \\
		\hline
		CDM & - &	- &	- &	- &	$75.1\pm2.8$ &	$75.2\pm2.8$ &	$75.3\pm2.7$ &	$75.2\pm2.7$ \\
		\hline
		SHFA &  $67.8\pm0.7$ &	$68.1\pm0.6$ &	$68.4\pm0.7$ &	$68.0\pm0.7$ &	$74.1\pm0.4$ &	$74.5\pm0.6$ &	$74.7\pm0.5$ &	$74.5\pm0.5$\\
		\hline
		G-JDA &  $69.4\pm0.8$ &	$70.5\pm0.7$ &	$69.6\pm1.0$ &	$70.1\pm1.0$ &	$76.0\pm0.7$ &	$76.8\pm0.8$ &	$76.8\pm0.7$ &	$76.6\pm0.7$\\
		\hline
		TNT &  $70.0\pm1.0$ &	$70.9\pm0.8$ &	$69.9\pm0.9$ &	$71.0\pm0.9$ &	$76.1\pm0.7$ &	$77.5\pm0.8$ &	$77.1\pm1.1$ &	$76.8\pm0.9$\\
		\hline
		CDLS &  $70.8\pm0.7$ &	$71.2\pm0.8$ &	$71.0\pm0.8$ &	$71.7\pm0.8$ &	$77.4\pm0.7$ &	$77.6\pm0.7$ &	$78\pm0.6$	& $77.5\pm0.7$\\
		\hline
		HDANA &  $72.1\pm0.3$ &	$72.7\pm0.4$ &	$71.8\pm0.4$ &	$72.3\pm0.3$ &	$78.3\pm0.5$ &	$79.2\pm0.5$ &	$80.1\pm0.6$ &	$78.5\pm0.3$\\
		\hline \hline
		HANDA (No SDL)& $23.4\pm8.2$ &	$16.8\pm9.0$ & $21.3\pm9.8$ &	$18.0\pm6.7$ &	$29.1\pm12.4$ &	$34.0\pm8.9$ &	$39.5\pm8.8$ &	$32.0\pm10.9$\\
		\hline
		HANDA (No Adv)&  $68.0\pm0.6$ &	$69.1\pm0.6$ &	$67.9\pm0.8$ &	$69.8\pm0.7$ &	$75.7\pm0.5$ &	$75.4\pm0.5$ &	$75.0\pm0.6$ &	$76.7\pm0.6$\\
		\hline
		HANDA (Entropy)&  $73.7\pm0.3$ &	$72.6\pm0.5$ &	$73.4\pm0.2$ &	$72.4\pm0.5$ &	$79.0\pm0.4$ &	$79.1\pm0.3$ &	$79.5\pm0.4$ &	$78.1\pm0.2$\\
		\hline
		\textbf{HANDA (ours)} & \bm{$74.5}\pm\bm{0.4^{\ast}$} &	\bm{$74.2}\pm\bm{0.4^{\ast}$} & \bm{$73.8}\pm\bm{0.5^{\ast}$} &	\bm{$74.0}\pm\bm{0.6^{\ast}$} &	\bm{$79.8}\pm\bm{0.3^{\ast}$} &	\bm{$80.0}\pm\bm{0.4^{\ast}$} &	\bm{$80.5}\pm\bm{0.2^{\ast}$} &	\bm{$79.2}\pm\bm{0.5^{\ast}$}\\
		\hline		
	\end{tabular}
\end{table*}

\selectlanguage{english}
\subsection{Experiment \#1: {Performance Evaluation}}
Experiment \#1 aims to evaluate HANDA’s performance against the state-of-the-art HDA methods in two relevant applications in international e-commerce platforms: multilingual text classification and cross-domain product image recognition. For multilingual text classification, following the settings in \cite{wang_heterogeneous_2018}, we summarize the results of domain adaptation when Spanish is the target and English, French, German, and Italian are the source languages, with 10 and 20 labeled target samples per class (Table \ref{table_result1}). The average accuracy is obtained by running each experiment 10 times and precedes the average standard deviation in Table \ref{table_result1} (separated by $\pm$). To see the effect of shared dictionary learning and adversarial kernel matching in isolation, we also conducted two baseline experiments with eliminating the shared dictionary learning component (No SDL) and eliminating the adversarial kernel matching component (No Adv). We further compared the performance of the proposed adversarial kernel matching alignment loss in HANDA with the entropy-based alignment loss proposed in \cite{morerio2018minimalentropy}, denoted by HANDA (Entropy) in Table {\ref{table_result1}}. $\beta$ and $\gamma$ were obtained via a small gird search as described in Experiment \#1.1. The higher performance of the proposed method is statistically significant when compared to the second-best method, as suggested by the paired $t$-test \cite{demsar_statistical_2006}. HANDA improves the classification performance by approximately 2\% across different source-target language pairs with statistically significant margins. Additionally, the baseline experiments show that both shared dictionary learning and adversarial kernel matching contribute to achieving the state-of-the-art performance. For cross-domain product image recognition, consistent with \cite{wang_heterogeneous_2018}, due to the lack of labeled data in the DSLR domain, we choose the Amazon (A), Webcam(W), and Caltech(C) domains as the source, and DSLR (D) as the target domain. HANDA improves the average accuracy in all domains by 1.1\% on average (Table \ref{table_result2}).
\begin{table}[!h]
	\setlength\abovecaptionskip{-0.3\baselineskip}
	\centering
	\caption{Cross-Features and Cross-Domain Comparison of State-of-the-Art HDA Methods with HANDA on Product Image Recognition; P-values significant at 0.05:*}
	\label{table_result2}
	\begin{tabular}{|M{0.15\linewidth}|M{0.2\linewidth}|M{0.2\linewidth}|M{0.2\linewidth}|M{0.16\linewidth}|}
		\hline
		HDA & \multicolumn{4}{c|}{SURF to DeCAF} \\\cline{2-5}
		
		Method & $A\rightarrow D$ & $C\rightarrow D$ & $W\rightarrow D$ & $Average$ \\
		\hline
		MMDT & $90.5\pm0.6$	& $91.2\pm0.6$ &	$90.8\pm0.6$ &	$90.8\pm0.6$ \\
		\hline
		TNT & $90.6\pm0.9$ &	$92.7\pm1.1$ &	$95.5\pm1.0$ &	$92.9\pm1.0$	 \\
		\hline
		SHFA & $93.4\pm1.1$	& $93.8\pm1.0$ &	$92.4\pm0.9$ &	$93.2\pm1.0$\\
		\hline
		G-JDA & $94.3\pm0.7$ &	$92.8\pm0.8$ 	&	$95.0\pm0.4$ 	&	$94.0\pm0.6$\\
		\hline
		CDLS &  $96.1\pm0.7$ &	$94.9\pm1.5$ &	$95.1\pm0.8$ &	$95.4\pm1.5$ \\
		\hline
		HDANA &  $96.1\pm0.5$ &	$95.3\pm0.6$ &	$96.9\pm0.3$ &	$96.1\pm0.5$\\
		\hline\hline
		HANDA \par (No SDL) &  $20.6\pm4.5$ &	$24.2\pm3.8$ &	$26.1\pm5.1$ &	$23.6\pm4.5$\\
		\hline
		HANDA \par (No Adv) &  $92.8\pm0.5$ &	$93.2\pm0.6$ &	$93.3\pm0.5$ &	$93.1\pm0.5$\\
		\hline
		HANDA \par(Entropy) &  $96.5\pm0.6$ &	$95.8\pm0.4$ &	$96.6\pm0.5$ &	$96.3\pm0.5$\\
		\hline
		\textbf{HANDA (ours)} & \bm{$97.1}\pm\bm{0.4^{\ast}$} & \bm{$96.9}\pm\bm{0.5 ^{\ast}$} & \bm{$97.6}\pm\bm{0.5^{\ast}$} &	\textbf{97.2}$\pm$\textbf{0.4}$\bm{^{\ast}}$  \\
		\hline	
	\end{tabular}
\end{table}
Overall, HANDA outperforms mathematical optimization-based HDA methods in text classification and product image recognition (the top right quadrant in Table \ref{table_positioning}). Moreover, HANDA outperforms the neural representation-based HDA alternatives (the bottom right quadrant in \ref{table_positioning}) in both text classification and product image recognition tasks. Outperforming HDANA suggests that domain invariance through adversarial learning can lead to better HDA. Outperforming TNT, in the image classification task by a significant margin suggests that simultaneous distribution and feature alignment are necessary for successful HDA.

\selectlanguage{english}
\subsubsection{Experiment \#1.1: Sensitivity to Parameters $\beta$ and $\gamma$}
\label{sensitivity}
As noted in Section 4, to empirically search the parameter space induced by $\beta$ and $\gamma$ in {({\ref{eq_14}})}, we conducted a grid search with four $\beta$ values selected from \{$1e-2,1e-3,1e-4,1e-5$\} and four $\gamma$ values selected from \{$1e-2,1e-1,1,10$\}. Following \cite{Ganin2016}, reverse cross-validation \cite{zhong2010cross} is adopted for hyperparameter tuning to ensure that validation is not conducted on labeled target data. Table {\ref{table_gridsearch}} shows the parameter values associated with the best performance during the empirical parameter tuning for $\beta$ and $\gamma$ in {({\ref{eq_14}})} on multilingual text dataset (with 10 and 20 labeled data in the source domain) and product image dataset (with three labeled data in the source domain).

\begin{table}[H]
	\setlength\abovecaptionskip{-0.3\baselineskip}
	\renewcommand{\arraystretch}{1.2}
	\caption{Results of Empirical Parameter Tuning for $\beta$ and $\gamma$ in {(\ref{eq_14})} on  Multilingual Text Dataset and Image Product Recognition Dataset}
	\label{table_gridsearch}
	\centering
	\begin{tabular}{|M{0.2\linewidth}|M{0.15\linewidth}|M{0.15\linewidth}|M{0.15\linewidth}|M{0.15\linewidth}|}
		\hline
		\multirow{2}{*}{HDA Task}& \multicolumn{2}{c|}{10 Labels / 3 Labels} & \multicolumn{2}{c|}{20 Labels}\\\cline{2-5}
		& $\beta (SDL)$ & $\gamma (Adv)$ & $\beta (SDL)$  & $\gamma (Adv)$ \\
		\hline
		$EN\rightarrow SP$ & $1e-5$ & $1$ & $1e-2$ &	$1e-4$   \\
		\hline
		$FR\rightarrow SP$ & $1e-4$ &	$1$ & $1e-5$ &	$1$ \\
		\hline
		$GR\rightarrow SP$ & $1e-4$ &	$1$ &	$1e-1$ &	$1e-4$ \\
		\hline
		$IT\rightarrow SP$ &  $1e-4$ & $1$ &	$1e-4$ &	$1$ \\
		\hline				
		$A\rightarrow D$ &  $1e-1$ &	$1e-4$ &	$-$ &	$-$ \\
		\hline
		$C\rightarrow D$ &  $1e-4$ &	$1$ &	$-$ &	$-$ \\
		\hline
		$W\rightarrow D$ &  $1e-4$ &	$1$ &	$-$ &	$-$ \\
		\hline
	\end{tabular}
\end{table}

As seen, a sparse search in the parameter space induced by $\beta$ and $\gamma$ yields the state-of-the-art performance reported in Table {\ref{table_result1}}. Additionally, the fact that $\beta=1e-4$ and $\gamma=1$ yields the best performance in the majority of domain adaptation experiments for both benchmark datasets signifies that {({\ref{eq_14}})} is not unreasonably sensitive to parameter setting.

\selectlanguage{english}
\subsubsection{Experiment \#1.2: Ablation analysis}
To further analyze the effect of the internal components of HANDA architecture, we conducted two sets of ablation experiments. The first set examines the effect of the depth of the feature extractor. Best performances are shown in bold face. As seen in Table \ref{table_ablation}, the results show performance improvement with multiple hidden layers for both datasets.
\begin{table*}[b]
	\setlength\abovecaptionskip{-0.3\baselineskip}
	\centering
	\caption{Ablation Analysis of the Depth of Feature Extractor in Multilingual Text Classification Dataset and Product Image Recognition Dataset}
	\label{table_ablation}
	\begin{tabular}{|M{0.17\linewidth}|M{0.1\linewidth}|M{0.1\linewidth}|M{0.1\linewidth}|M{0.1\linewidth}|M{0.1\linewidth}|M{0.1\linewidth}|M{0.1\linewidth}|M{0.1\linewidth}|}
		\hline
		\multirow{2}{*}{Model}& \multicolumn{4}{c|}{10 Labeled Samples per Target Class} & \multicolumn{4}{c|}{20 Labeled Samples per Target Class}\\\cline{2-9}
		& $EN\rightarrow SP$ & $FR\rightarrow SP$ & $GR\rightarrow SP$ & $IT\rightarrow SP$  & $EN\rightarrow SP$ & $FR\rightarrow SP$ & $GR\rightarrow SP$ & $IT\rightarrow SP$\\
		\hline
		1-layer & $73.0\pm0.4$ & $72.7\pm0.5$ & $71.6\pm0.6$ &	$72.2\pm0.5$ & $77.5\pm0.2$ &	$77.9\pm0.5$ & $79.6\pm0.3$ &	$78.4\pm0.4$  \\
		\hline
		2-layer & $\bm{74.5}\pm\bm{0.4}$ & $74.2\pm0.4$ & $\bm{73.8}\pm\bm{0.5}$ & $74.0\pm0.6$ & $79.8\pm0.3$ & $\bm{80.0}\pm\bm{0.4}$ &	$\bm{80.5}\pm\bm{0.2}$ &	$\bm{79.2}\pm\bm{0.5}$ \\
		\hline
		3-layer &  $74.4\pm0.3$ & $\bm{74.7}\pm\bm{0.6}$ &	$72.2\pm0.7$ &	$\bm{74.3}\pm\bm{0.7}$ & $80.0\pm0.4$ &	$79.3\pm0.3$ &	$80.1\pm0.4$ &	$79.1\pm0.5$\\
		\hline
		4-layer &  $74.2\pm0.5$ & $74.0\pm0.5$ &	$72.5\pm0.5$ &	$73.5\pm0.6$ &	$\bm{80.2}\pm\bm{0.4}$ &	$79.3\pm0.4$ &	$79.5\pm0.3$ &	$78.3\pm0.6$\\
		\hline
		5-layer &  $72.3\pm0.7$ & $72.5\pm0.7$ &	$72.0\pm0.8$ &	$72.1\pm0.9$ &	$79.5\pm0.5$ &	$78.6\pm0.5$ &	$78.2\pm0.5$ &	$77.9\pm0.6$\\
		\hline
		SDL+Adv (Sequential) & $71.1\pm0.4$ & $71.2\pm0.5$ & $70.4\pm0.5$ & $69.9\pm0.6$ & $76.6\pm0.4$ & $76.2\pm0.5$ & $75.8\pm0.5$ &	$75.2\pm0.5$ \\
		\hline	
	\end{tabular}
\end{table*}
It is observed that having three (and rarely four) hidden layers adds marginal improvements compared to two hidden layers. That is, the majority of best performances are attained with two hidden layers. As such, to promote a parsimonious architecture that yields the best performance in most heterogeneous domain adaptation tasks, we used a neural net with two hidden layers as the feature extractor of HANDA. The second set of experiments examines the contribution of simultaneous learning of the shared dictionary and adversarial kernel matching in the HANDA's unified framework versus the sequential learning of these two components. In the sequential case, denoted by (SDL+Adv) in Table \ref{table_ablation}, we conducted a two-step learning process, in which we first optimized HANDA's shared dictionary learning loss (without involving adversarial kernel matching loss). Subsequently, we applied the adversarial kernel learning on the projected space obtained from the shared dictionary learning. The results suggest that simultaneous learning of the shared dictionary and adversarial kernel matching in HANDA's unified architecture outperforms sequential learning on multilingual text classification dataset.

\selectlanguage{english}
\subsection {Experiment \#2: Qualitative Analysis}
\begin{figure}[!h]

	\subfloat[]{\includegraphics[width=0.2\textwidth]{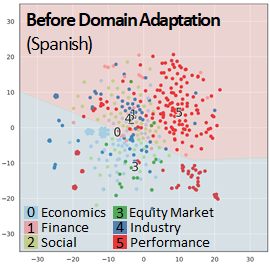}%
		\label{fig_svm_lang_a}}
	\hfil
	\subfloat[]{\includegraphics[width=0.2\textwidth]{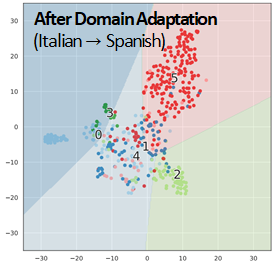}%
		\label{fig_svm_lang_b}}
	\hfil
	\subfloat[]{\includegraphics[width=0.2\textwidth]{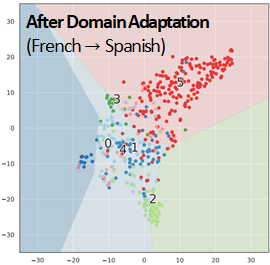}%
		\label{fig_svm_lang_c}}
	\hfil
	\subfloat[]{\includegraphics[width=0.2\textwidth]{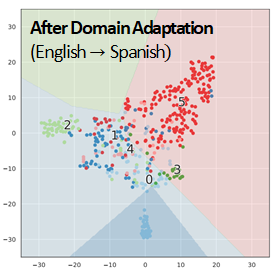}%
		\label{fig_svm_lang_d}}
	\caption{Qualitative Comparison of HANDA Before and After Domain Adaptationin in Multilingual Text Classification. Target samples have been visualized in 2-D before and after domain adaptation from Italian (b), French (c), and English (d) to Spanish.}
	
	\label{fig_svm_lang}
\end{figure}

\begin{figure}[!h]
	\centering
	\setlength\abovecaptionskip{0.1\baselineskip}	\setlength\belowcaptionskip{-0.5\baselineskip}
	\setlength\dbltextfloatsep{-0.5\baselineskip}
	\subfloat[]{\includegraphics[width=0.2\textwidth]{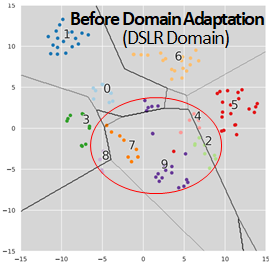}%
		\label{fig_svm_image_a}}
	\hfil
	\subfloat[]{\includegraphics[width=0.2\textwidth]{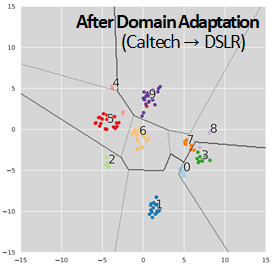}%
		\label{fig_svm_image_b}}
	\hfil
	\captionsetup[subfigure]{labelformat=empty}
	\subfloat[]{\includegraphics[width=0.25\textwidth]{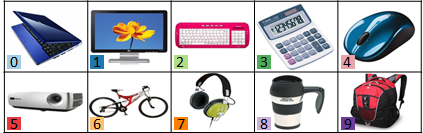}%
		\label{fig_svm_image_legend}}
	\hfil	
	\vspace{-10pt}
	\caption{Qualitative Comparison of HANDA Before and After Domain Adaptation in Cross-Domain Product Image Recognition. Target samples are visualized in 2-D before (a) and after (b) domain adaptationfrom Caltech to DSLR domain.}
	\label{fig_svm_image}
\end{figure}

In domain adaptation research, the quality of generated representations can be assessed via visualizing the obtained representations before and after DA \cite{chen_transfer_2016}. The target representation from HANDA is visualized in 2-D space by $t$-distributed Stochastic Neighbor Embedding ($t$-SNE) \cite{maaten_visualizing_2008} (Figure \ref{fig_svm_lang}). The decision boundaries are obtained by linear SVM.
As seen in Figure \ref{fig_svm_lang_a}, target documents are not linearly separable before DA. The new representations obtained from HANDA lead to more linearly separable samples after domain adaptation from Italian (Figure \ref{fig_svm_lang_b}), French (Figure \ref{fig_svm_lang_b}), and English (Figure \ref{fig_svm_lang_d}). The majority of documents in the ‘Economics,’ ‘Social,’ and ‘Performance’ classes are distinguished via light blue, light green, and dark red hyperplanes, respectively. The same effect is observed in the cross-domain product image recognition task as a result of DA (Figure \ref{fig_svm_image}).
Similarly, target images are not linearly separable before DA (Figure \ref{fig_svm_image_a}). A considerable number of product images in ‘mouse’ (light red), ‘headphone’ (orange), ‘backpack’ (dark purple), ‘mug’ (light purple), and ‘keyboard’ (light green) cannot be distinguished (shown with a red circle). HANDA representations lead to more linearly separable samples after adaptation from Caltech to DSLR (Figure \ref{fig_svm_image_b}) compared to before adaptation (Figure \ref{fig_svm_image_a}). Almost all product images are classified correctly except the ‘laptop’ (light blue) and ‘calculator’ (dark green) that have excessive visual similarity (both have a display and keypad).
\selectlanguage{english}
\subsection{Experiment \#3: Convergence Analysis}
\begin{figure}[!h]
	\setlength\abovecaptionskip{0.2\baselineskip}	\setlength\belowcaptionskip{-0.5\baselineskip}
	\setlength\dbltextfloatsep{-0.5\baselineskip}
	\centering
	\subfloat[]{\includegraphics[width=0.31\linewidth]{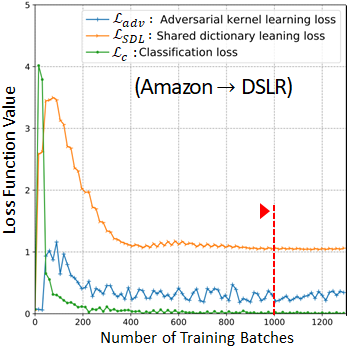}%
		\label{fig_loss_a}}
	\hfil
	\subfloat[]{\includegraphics[width=0.3\linewidth]{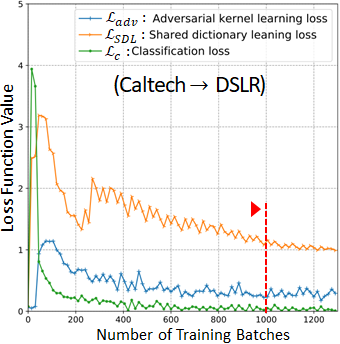}%
		\label{fig_loss_b}}
	\hfil
	\subfloat[]{\includegraphics[width=0.3\linewidth]{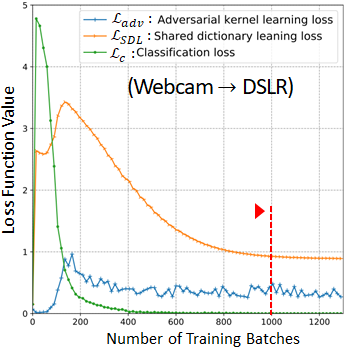}%
		\label{fig_loss_c}}
	\hfil
	
	\caption{Convergence Analysis of HANDA on Amazon (a), Caltech (b), and Webcam (c) Domains for Image Recognition Task on Office31-Caltech256 Dataset. Loss variations stabilize after a certain number of iterations in all domains.}
	\label{fig_loss}
\end{figure}

The quality of adversarial training is often empirically verified by investigating the convergence property of the losses \cite{arjovsky_wasserstein_2017}. We monitor adversarial kernel learning, shared dictionary learning, and classification loss on Office31 – Caltech256. All three losses stabilize after 1,000 training batches (red markers in Figure \ref{fig_loss}). Although both dictionary learning and adversarial kernel learning have non-convex objective functions, they both converge in all three domains (Amazon (Figure \ref{fig_loss_a}), Caltech (Figure \ref{fig_loss_b}), and Webcam (Figure \ref{fig_loss_c})).
In sum, with our feature and distribution alignment approaches, HANDA is able to outperform extant HDA methods in image recognition and text classification tasks. Specifically, outperforming HANDA alternatives, HDANA and TNT, shows that simultaneous feature and distribution alignment with a focus on domain invariance can lead to better HDA. HANDA can alleviate the lack of training data in e-commerce applications and improve product search and indexing in online markets. Furthermore, HANDA shows loss convergence property during the adversarial training process. This suggests that HANDA can be robust and generalizable to unknown domains. Finally, the domain-invariant representations obtained from HANDA are of high quality and can increase the performance of downstream text and image classification tasks.
\selectlanguage{english}
\subsection{Cybersecurity Case Study}
\begin{figure}[!h]
	\centering
	\setlength\abovecaptionskip{0.2\baselineskip}	\setlength\belowcaptionskip{-0.5\baselineskip}
	\setlength\dbltextfloatsep{-0.5\baselineskip}
	\includegraphics[scale=0.5]{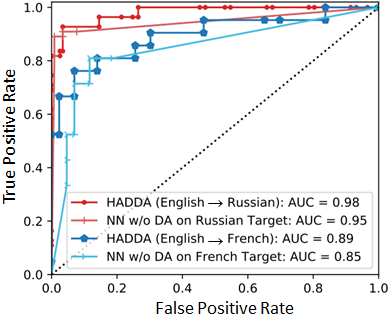}
	\caption{HANDA’s AUC on Russian and French Dark Web Online Markets.
		Note: Softmax function was applied to all network outputs to obtain the probabilities for AUC calculation.
	}
	\label{fig_auc}
\end{figure}

It is useful to assess the utility of HDA models in real-world applications, in addition to common benchmark evaluations. Monitoring products sold on dark web markets is an emerging area in cybersecurity that can highly benefit from HDA in providing actionable cyber threat intelligence. We demonstrate the practical utility of the HANDA framework on a real-world cybersecurity dataset obtained from English, Russian, and French dark web markets. This case study has two main purposes. First, it aims to assess if a model that is trained by Algorithm 1 on English markets can aid the detection of cyber threats in Russian and French markets in practice. Second, it is meant to show examples of cyber threats that are recognized by HANDA but would be missed in the absence of DA. To gauge the performance improvement offered by HANDA, we compare the AUC to a neural network without DA (Figure \ref{fig_auc}).
To this end, a feedforward neural network (NN) trained on Russian and French markets serves as a baseline. HANDA was trained on English markets as the source, while Russian and French were the targets. HANDA increases the AUC in Russian markets by 3\% (from 0.96 to 0.98). Similarly, HANDA results in a 4\% AUC increase in French markets (from 0.85 to 0.89). Table \ref{table_threat} shows the examples of products that HANDA identifies in Russian and French as cyber threats, but are missed by the baseline NN without DA. By juxtaposing the training samples, testing samples, and model’s output, we discovered that there were no instances of “Trojan” in the French training set. Nevertheless, HANDA was able to identify the Trojan as a cyber threat in the French testing set, showing that our model is able to learn from Trojan instances in the source (English) dataset. This implies that the domain adaptation from English to French allows HANDA to distinguish the ‘unseen’ examples in the target language that are not present in the training set. As a result, HANDA can significantly increase the cyber threat intelligence performance in dark web markets with limited training data.

\begin{table*}[t]
	\setlength\abovecaptionskip{-0.2\baselineskip}
	\centering
	\caption{Russian and French Product Descriptions Correctly Identified as Cyber Threats by HANDA while Missed by the Baseline Method.}
	\label{table_threat}
	\begin{tabular}{|M{0.1\linewidth}|M{0.07\linewidth}|M{0.27\linewidth}|M{0.28\linewidth}|M{0.09\linewidth}|M{0.08\linewidth}|}
		\hline
		& Language & Product Description Excerpt & Translation by Native Speaker & Cyber Threat Category & HANDA Confidence Score\\
		\hline
		\raisebox{-.3\height}{\includegraphics[scale=.9]{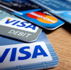}} & Russian & ‘‘\begin{otherlanguage*}{russian}банковской карты без физического носителя.После покупки вы получите данные в следующем формате:[…]. \end{otherlanguage*}’’ & Bank card data without physical media. After the purchase, you will receive data in the following format: […]. &  Stolen financial credentials & 0.9\\
		\hline
		\raisebox{-.3\height}{\includegraphics[scale=1]{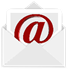}} & Russian &	‘‘\begin{otherlanguage*}{russian}Взлом GMAIL.COM-Клиент сообщает предмет взлома|ID жертвы и.т.п[…] После взлома, клиент получает доказательства проделанной работы.-Оплата заказа […].\end{otherlanguage*}’’  &	Hacking GMAIL.COM-The client reports the subject of hacking | victim ID, etc.[…]-After hacking, the client receives evidence of the work done […]. &	E-mail hacking service & 0.9 \\
		\hline
		\raisebox{-.3\height}{\includegraphics[scale=1]{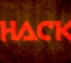}} & French &	 ‘‘\begin{otherlanguage*}{french}Ce logiciel vous permet de checker vos logs pour trouver ceux qui ont un accés à la boite email. Une fois un email:pass valide trouver une option vous permet de télécharger tous les documents de cette boite email […].\end{otherlanguage*}’’ &	This software allows you to check your logs to find those who have access to the mailbox. Once you have found a valid email: pass, there will be an option that will let you download all the documents from that mailbox […]. &	E-mail hacking tool &	0.7 \\
		\hline
		\raisebox{-.3\height}{\includegraphics[scale=0.9]{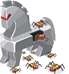}} & French &	 ‘‘\begin{otherlanguage*}{french}Un pack de 3 différents botnet rare avec 5 RAT et trojan, pour vos intrusions les plus importantes. Des centaines de pc victimes obéïssant à tous vos ordres […] .\end{otherlanguage*}’’ &	A pack of 3 different rare botnets with 5 RAT and Trojan, for your most important intrusions. Hundreds of pc victims obeying all your orders […]. &	Trojan &	0.6 \\
		\hline
	\end{tabular}
\end{table*}

\selectlanguage{english}
\section{Conclusion and Future Directions}
Given the growing adoption of machine learning models to analyze novel online markets, adapting previously learned models to unseen domains is a promising strategy. The adaptation process is challenging when dealing with heterogeneous domains where the source and target domains differ in both feature space and data distribution. This poses a significant problem in e-commerce and emerging fields such as cybersecurity. HDA is a crucial approach to address this emerging challenge. However, most extant HDA methods focus on minimizing the distance via non-neural representations, which may suffer from a lack of transferability. Adversarially learned representations have been shown to yield more transferrable representations. Nonetheless, they are not designed for HDA problems. To address this gap, we developed a novel framework for Heterogeneous Adversarial Neural Domain Adaptation (HANDA), a neural network architecture that employs dictionary learning and nonparametric adversarial kernel matching to jointly minimize the feature space discrepancy and distribution divergence in a unified architecture. HANDA extends adversarial domain adaptation for heterogeneous domains and incorporates shared dictionary learning in a neural network architecture to benefit from labeled data. We conducted in-depth evaluations to evaluate HANDA’s performance against state-of-the-art HDA methods in two benchmark learning tasks in common e-commerce applications (i.e., product image recognition and multilingual text classification). HANDA improves the classification performance on the benchmark datasets with statistically significant results. HANDA can be employed to improve product search and indexing in online markets. We also showed the advantage of utilizing HANDA in emerging dark web online markets. HANDA is able to better identify cyber threats among the products sold on dark web online markets. Two promising future directions are envisioned. First, improving the interpretability of the model by incorporating components such as attention mechanism can shed light on the DA process in HANDA. Second, given the prevalence of sequence data in e-commerce applications, extending HANDA to account for the temporal factor in sequential inputs is another promising area in HDA research.

% or


\selectlanguage{english}
\ifCLASSOPTIONcompsoc
  \section*{Acknowledgments}
\else
  \section*{Acknowledgment}
\fi

This material is based upon work supported by the National Science Foundation (NSF) under grants CNS-1936370 (SaTC CORE) and OAC-1917117 (CICI). Yidong Chai was supported in part by the NSFC (91846201, 72101079), Shanghai Data Exchange Cooperative Program (W2021JSZX0052), and is the corresponding author.

\ifCLASSOPTIONcaptionsoff
  \newpage
\fi



\selectlanguage{english}
\bibliographystyle{IEEEtran}
\bibliography{hda}
\selectlanguage{english}
\iffalse

\fi


%\vspace{-70pt}
\begin{IEEEbiography}[{\includegraphics[width=1in,height=1.25in,clip,keepaspectratio]{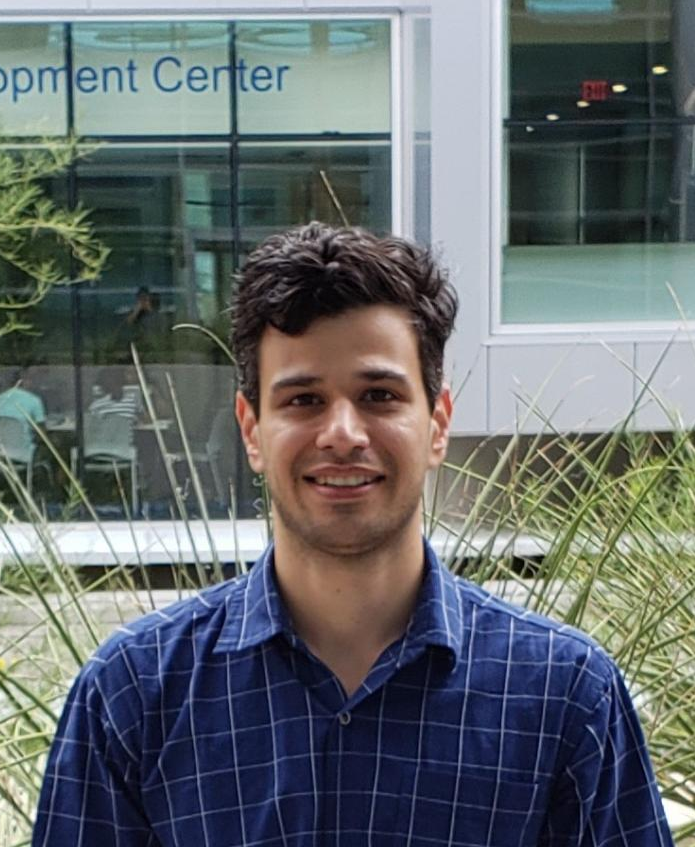}}]{Mohammadreza (Reza) Ebrahimi}
received his master’s degree in Computer Science from Concordia University, Canada in 2016, and his Ph.D. in Information Systems from Artificial Intelligence (AI) Lab at the University of Arizona in 2021. He is an assistant professor at the University of South Florida (USF). His research interests include statistical machine learning, adversarial machine learning, and security of AI. His work has appeared in journals, conferences, and workshops, including IEEE S\&P, AAAI, IEEE ISI, \textit{Digital Forensics}, and \textit{Applied Artificial Intelligence}. He has served as a Program Chair and Program Committee member in IEEE ICDM Workshop on Machine Learning for Cybersecurity (MLC) and IEEE S\&P Workshop on Deep Learning and Security (DLS). He has contributed to several projects supported by the National Science Foundation (NSF). He is a member of the IEEE, ACM, AAAI, and AIS.
\end{IEEEbiography}

\vspace{-20pt}
\begin{IEEEbiography}[{\includegraphics[width=1in,height=1.25in,clip,keepaspectratio]{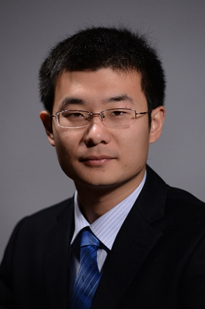}}]{Yidong Chai}
	received his bachelor’s degree in information system from Beijing Institute of Technology, and his Ph.D. degree in Management Information Systems from Tsinghua University. He is currently a professor at the Hefei University of Technology. His research fields include machine learning, signal processing, and natural language processing. His work has appeared in journals including Knowledge-Based Systems and Applied Soft Computing, as well as conferences and workshops including IEEE S\&P, INFORMS Workshop on Data Science, Workshop on Information Technology Systems, International Conference on Smart Health, and International Conference on Information Systems.
	
\end{IEEEbiography}

\vspace{-20pt}

\begin{IEEEbiography}[{\includegraphics[width=1in,height=1.25in,clip,keepaspectratio]{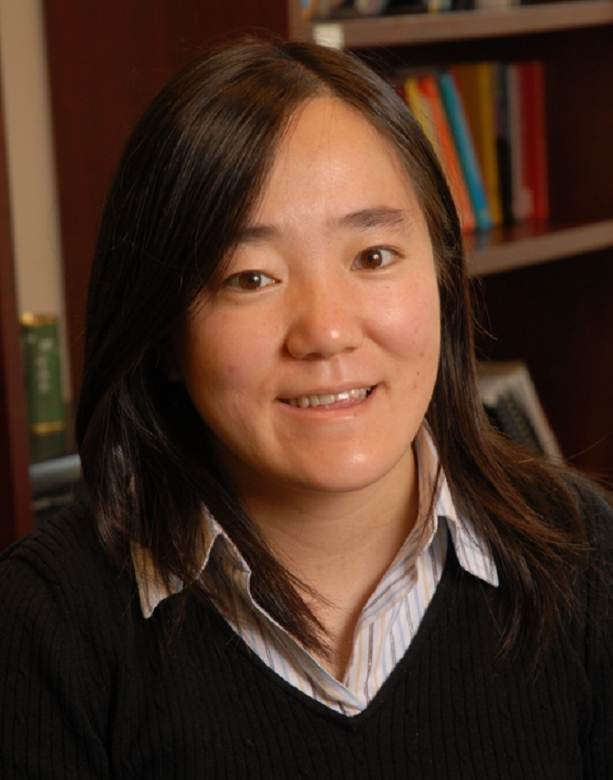}}]{Hao Helen Zhang}
	is a professor at the University of Arizona, in the Department of Mathematics, Statistics Interdisciplinary Program, and Applied Mathematics Interdisciplinary Program there. With Bertrand Clarke and Ernest Fokoué, she is the author of the book Principles and Theory for Data Mining and Machine Learning.	She earned a bachelor's degree in mathematics in 1996 from Peking University. She completed her Ph.D. in statistics in 2002 from the University of Wisconsin–Madison. Her dissertation, supervised by Grace Wahba, was Nonparametric Variable Selection and Model Building Via Likelihood Basis Pursuit. She was elected to the International Statistical Institute and as a fellow of the American Statistical Association in 2015. She became a fellow in the Institute of Mathematical Statistics in 2016, and has been selected as the 2019 Medallion Lecturer of the Institute of Mathematical Statistics.
\end{IEEEbiography}

\vspace{-20pt}

\begin{IEEEbiography}[{\includegraphics[width=1in,height=1.25in,clip,keepaspectratio]{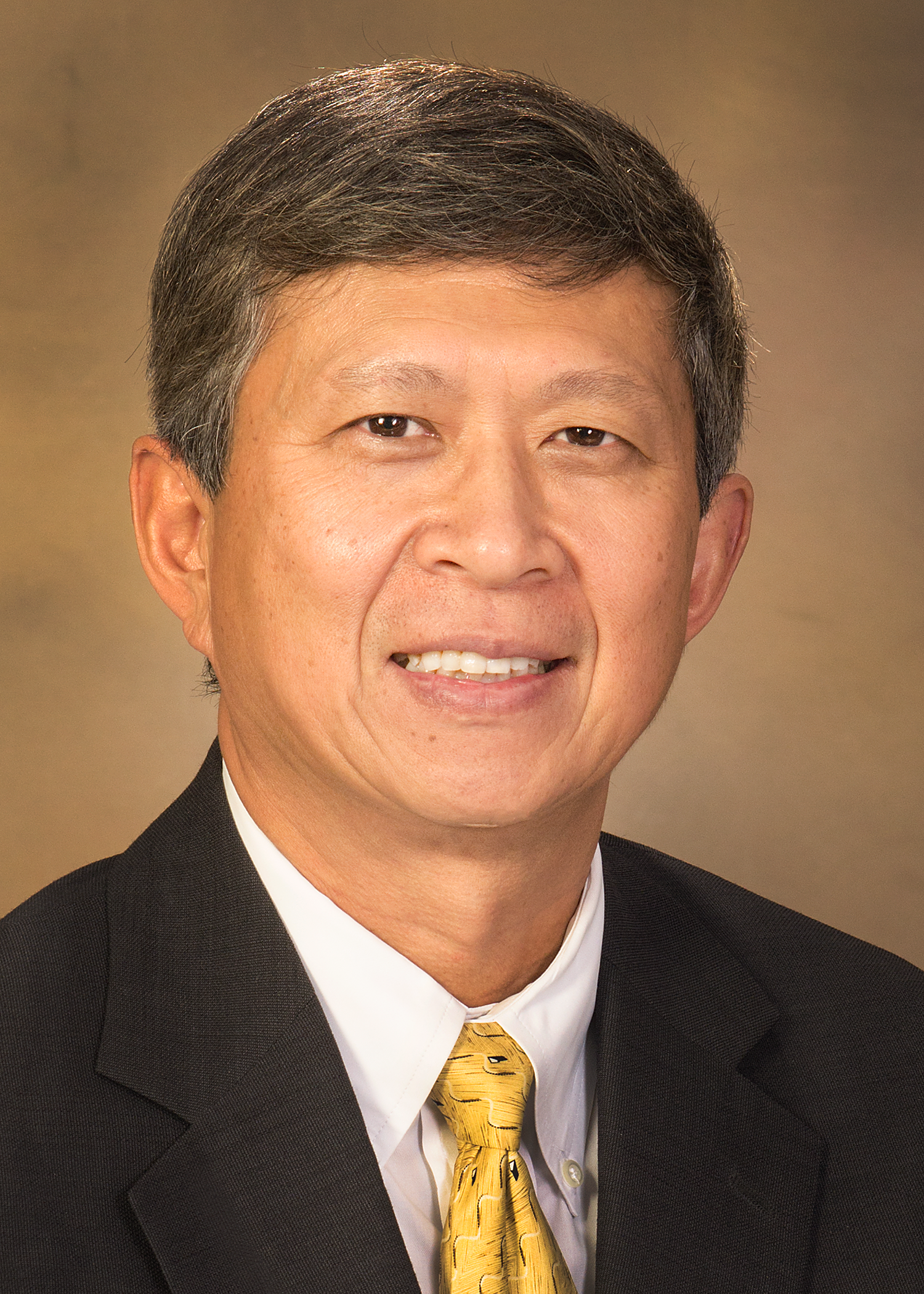}}]{Hsinchun Chen}
	received the BS degree from the National Chiao-Tung University in Taiwan, the MBA degree from the State University of New York at Buffalo, and the Ph.D. degree in information systems from New York University. He is a regents professor at the University of Arizona. He has served as a scientific counselor/advisor of the US National Library of Medicine, the Academia Sinica, Taiwan, and the National Library of China, China. He was ranked \#8 in publication productivity in information systems (CAIS 2005) and \#1 in Digital Library research (IP\&M 2005) in two bibliometric studies. His COPLINK system, which has been quoted as a national model for public safety information sharing and analysis, has been adopted in more than 550 law enforcement and intelligence agencies in 20 states. He received the IEEE Computer Society 2006 Technical Achievement Award. He is a fellow of the IEEE and the AAAS.
\end{IEEEbiography}

%\newpage


%\vfill


\end{document}